\documentclass{article} 
\usepackage{iclr2027_conference,times}

\usepackage{microtype}
\usepackage{graphicx}
\usepackage{subcaption}

\usepackage{amsmath,amsfonts,bm}

\def\eqref#1{equation~\ref{#1}}

\def\1{\bm{1}}

\DeclareMathAlphabet{\mathsfit}{\encodingdefault}{\sfdefault}{m}{sl}
\SetMathAlphabet{\mathsfit}{bold}{\encodingdefault}{\sfdefault}{bx}{n}

\newcommand{\sigmoid}{\sigma}

\DeclareMathOperator*{\argmin}{arg\,min}

\usepackage{amsmath}
\usepackage{amssymb}
\usepackage{mathtools}
\usepackage{amsthm}
\usepackage{bm}
\usepackage[most]{tcolorbox}
\usepackage{cancel}

\usepackage{hyperref}
\usepackage{url}
\usepackage{titletoc}
\usepackage{booktabs} 

\usepackage{wrapfig}
\usepackage{float}
\usepackage{enumitem}
\usepackage{circuitikz}
\usetikzlibrary{positioning,arrows.meta}

\usepackage{tabularx}
\usepackage{array}

\usepackage{xcolor}
\usepackage{empheq}
\newcommand{\highlightbox}[1]{%
    \tcboxmath[
        colframe=orange!35,
        colback=orange!7,
        boxrule=0.4pt,
        arc=2pt,
        left=3pt,
        right=3pt,
        top=3pt,
        bottom=3pt
    ]{#1}%
}
\hypersetup{
    colorlinks=true,
    linkcolor=blue!60!black,
    citecolor=blue!45!black,
    urlcolor=blue!60!black
}

\title{Analog-Friendly Predictive Coding without Activation Derivatives}

\author{Francesco Innocenti \\
Brain Network Dynamics Unit \\
University of Oxford, UK \\
\texttt{francesco.innocenti@ndcn.ox.ac.uk}
}

\iclrfinalcopy 
\begin{document}

\maketitle
\lhead{Preprint}
\begin{abstract}
    Predictive coding (PC) is a local, energy-based alternative to backpropagation (BP) whose iterative inference dynamics make it attractive for implementation on analog hardware. However, standard nonlinear PC requires evaluating the derivative of the activation function during both inference and learning, which can be difficult to realise physically. Here, we introduce \textit{activation-matched Bregman PC}, replacing standard squared-error energies with Bregman divergences matched to the activation function. This formulation eliminates activation derivatives and, when combined with inference via mirror descent, yields local inference and learning rules requiring only weighted sums, local prediction errors, state integration, and the activation function. In digital experiments, Bregman PC performs comparably to standard PC and BP on classification and generative tasks, while preserving characteristic learning dynamics of PC and its convergence to BP under stable large-model parameterisations. These results provide a more analog-friendly formulation of nonlinear PC while retaining its key computational properties.
\end{abstract}

\section{Introduction} \label{sec:intro}
As the energy demands of AI continue to grow \citep{de2023growing, luccioni2024power}, there is increasing interest in alternatives to conventional digital computing and backpropagation-based training \citep{hooker2021hardware}. Analog hardware can exploit the intrinsic dynamics of physical systems to perform inference with high energy efficiency \citep{shainline2021optoelectronic, wright2022deep, aifer2025solving, momeni2025training, wright2026physical, wang2026generative, lockwood2026blueprint}, even when model parameters are trained off-chip. Ultimately, however, further gains may come from performing learning directly on the physical substrate itself, reducing analog-to-digital communication and moving closer to the online adaptation of biological systems. This motivates learning algorithms built from local operations that can be realised directly in hardware.

Predictive coding (PC) is a promising but underexplored candidate for analog implementation. PC networks (PCNs) define local prediction errors between the activity of each layer and the prediction generated by the preceding layer, minimising an energy given by the sum of these errors \citep{rao1999predictive, friston2005theory}. During inference, neural activities evolve to reduce the energy. Once inference has converged, weights are updated using only locally available pre- and postsynaptic information. PC therefore replaces the explicit backward pass of backpropagation with local recurrent dynamics that can, in principle, be realised directly by a physical system.

Recent work on scaling PC to larger models and more complex tasks has been encouraging. Several studies have improved the training of deep PCNs \citep{qi2025towards, goemaere2025epc}, including networks with more than 100 layers \citep{innocenti2025mu, seely2026augmented}, and stable parameterisations have been derived for PCNs in large-width and depth limits \citep{ishikawa2024local, innocenti2026infinite}. More recently, \cite{kerjan2026training} showed that convolutional PCNs trained using equilibrium propagation can achieve BP-competitive performance on ImageNet classification.

Despite these promising advances, it remains largely unknown how PC could be implemented on analog hardware. One important obstacle is that standard nonlinear PC formulations require explicit evaluation of activation-function derivatives during both inference and learning (see Figure~\ref{fig:standard-vs-bregman}). While an analog device may naturally realise the activation function itself, separately computing its derivative can require additional circuitry or operations that are difficult or inefficient to implement on the same substrate. Motivated by biological plausibility, previous work has explored removing these activation derivatives from PC, but found that doing so significantly degraded performance on simple classification tasks \citep{millidge2020relaxing}. We therefore ask: \textbf{\textit{Can nonlinear PC be formulated without explicit activation derivatives, making it better suited to analog hardware, while preserving its key computational properties?}} 

Here, we provide a positive answer to this question. Building on previous activation-matched Bregman constructions \citep{wang2020generalised, amid2022locoprop, wang2023lifted}, we show that activation derivatives can be eliminated by replacing the squared Euclidean prediction errors of standard PC with Bregman divergences whose geometry is matched to the activation function. 
Combined with mirror descent of the neural states, we obtain local inference and learning updates that require only weighted interactions, local prediction errors, state integration, and evaluation of the activation itself (Figure~\ref{fig:standard-vs-bregman}). The resulting formulation recovers standard PC for linear networks while providing nonlinear PC dynamics that are more naturally suited to analog implementation.

The rest of the paper is organised as follows. We first review standard PC and Bregman divergences (\S\ref{sec:background}), highlighting the origin of activation derivatives in nonlinear PC. In \S\ref{sec:bregman-pc}, we introduce activation-matched Bregman-PC, deriving its energy, mirror-descent inference dynamics, and local learning rule. We then evaluate Bregman-PC on standard learning tasks and test whether it preserves key computational properties of PC (\S\ref{sec:exps}). We conclude by discussing the main limitations and future directions of this work (\S\ref{sec:discussion}).
\begin{figure}[t]
    \centering
    \resizebox{\linewidth}{!}{%
        \begin{tikzpicture}[
            font=\small,
            cell/.style={
                draw=black!25,
                fill=black!2,
                rounded corners=2pt,
                align=center,
                text width=6.7cm,
                minimum height=1.3cm,
                inner sep=5pt
            },
            stdcell/.style={
                cell,
                draw=black!25,
                fill=black!2
            },
            bregcell/.style={
                cell,
                draw=orange!35,
                fill=orange!7
            },
            rowlabel/.style={
                font=\bfseries,
                align=left,
                text width=1.3cm
            },
            header/.style={
                font=\bfseries,
                align=center
            }
        ]
        
        \node[header, text=black!70] (std-header) at (0,0)
            {Standard predictive coding};
        
        \node[header, text=orange!70!black] (breg-header) at (7.2,0)
            {Bregman predictive coding};
        
        \node[stdcell] (std-error) at (0,-1.15) {%
            \(\displaystyle \mathcal{F}^\ell = \frac{1}{2} \big\| \bm{\varepsilon}^\ell \big\|_2^2 = \frac{1}{2} \big\| \bm{z}^\ell-\phi(\bm{a}^\ell) \big\|_2^2\)
        };
        
        \node[bregcell] (breg-error) at (7.2,-1.15) {%
            \(\displaystyle \mathcal{F}^\ell_{\mathrm B} = D_\Psi\left( \bm{z}^\ell,\phi(\bm{a}^\ell) \right), \quad \nabla\Psi(\bm{z}^\ell)=\phi^{-1}(\bm{z}^\ell)\)
        };
        
        \node[rowlabel, left=4mm of std-error] {Error measure};

        \node[stdcell] (std-inference) at (0,-2.75) {%
            \(\displaystyle \frac{d\bm{z}^\ell}{d\tau} = -\bm{\varepsilon}^\ell + \textcolor{orange!75!black}{\phi'(\bm{a}^{\ell+1})} \odot (\bm{W}^{\ell+1})^\top \bm{\varepsilon}^{\ell+1}\)
        };
        
        \node[
            bregcell,
        ] (breg-inference) at (7.2,-2.75) {%
            \(\displaystyle \frac{d\bm{u}^\ell}{d\tau} = - \bm{u}^\ell + \bm{a}^\ell + (\bm{W}^{\ell+1})^\top \bm{\varepsilon}^{\ell+1}, \quad \bm{z}^\ell = \phi(\bm{u}^\ell)\)
        };
        
        \node[rowlabel, left=4mm of std-inference] {Inference};

        \node[stdcell] (std-learning) at (0,-4.35) {%
            \(\displaystyle \frac{d\bm{W}^\ell}{dt} = \eta\, \textcolor{orange!75!black}{\phi'(\bm{a}^\ell)} \odot \bm{\varepsilon}^\ell (\bm{z}^{\ell-1})^\top\)
        };
        
        \node[bregcell] (breg-learning) at (7.2,-4.35) {%
            \(\displaystyle \frac{d\bm{W}^\ell}{dt} = \eta\, \bm{\varepsilon}^\ell (\bm{z}^{\ell-1})^\top\)
        };
        
        \node[rowlabel, left=4mm of std-learning] {Learning};
        
        \end{tikzpicture}
    }
    \caption{\textbf{Standard predictive coding versus activation-matched Bregman predictive coding.} Both formulations use the same nonlinear prediction \(\widehat{\bm{z}}^\ell=\phi(\bm{a}^\ell)\) (see Table~\ref{tab:key-notation} for notation), but differ in the error measure used at each hidden layer. Standard PC uses a squared Euclidean error (Eq.~\ref{eq:pc-energy}), resulting in explicit activation derivatives \(\textcolor{orange!75!black}{\phi'}\) in both inference and learning (Eq.~\ref{eq:pc-inference} \& Eq.~\ref{eq:pc-learning}). Bregman PC instead uses an activation-matched Bregman divergence (Eq.~\ref{eq:bregman-energy}) satisfying \(\nabla\Psi(\bm{z}^\ell)=\phi^{-1}(\bm{z}^\ell)\). This removes the explicit activation derivative: inference evolves the dual state \(\bm{u}^\ell\) (Eq.~\ref{eq:bregman-inference}), with \(\bm{z}^\ell=\phi(\bm{u}^\ell)\), while learning reduces to a local error-activity outer product (Eq.~\ref{eq:bregman-weight-update}).}
    \label{fig:standard-vs-bregman}
\end{figure}
\subsection{Summary of contributions}
\begin{itemize}
    \item \textbf{Activation-matched Bregman PC without explicit activation derivatives.}
    We formulate nonlinear PC using activation-matched Bregman divergences as layerwise prediction errors and mirror descent in the corresponding dual coordinates. The resulting local inference and learning rules do not require explicit activation derivatives (Figure~\ref{fig:standard-vs-bregman}), while monotonically decreasing the Bregman-PC energy and recovering standard PC exactly in the linear case.
    \item \textbf{Empirical validation on simple learning tasks.} Using digital simulations on classification and generative learning tasks, we show that Bregman PC trains nonlinear networks with performance comparable to standard PC and BP (Figure~\ref{tab:benchmark-performance}), demonstrating that removing explicit activation derivatives does not prevent effective learning in practice.
    \item \textbf{Preservation of established PC learning regimes.} We show that Bregman PC retains learning regimes previously established for standard PC (Figure~\ref{fig:pc-regimes}), including fast ``saddle-to-saddle dynamics'' near degenerate saddles and convergence of PC gradients to BP under stable large-model parameterisations.
\end{itemize}

\section{Background} 
\label{sec:background}
\begin{table}[t]
    \centering
    \caption{Key notation.}
    \label{tab:key-notation}
    \small
    \setlength{\tabcolsep}{4pt}
    \begin{tabularx}{\linewidth}{
        @{}
        >{\raggedright\arraybackslash}p{0.12\linewidth}
        >{\raggedright\arraybackslash}p{0.27\linewidth}
        >{\raggedright\arraybackslash}X
        @{}
    }
    \toprule
    \textbf{Symbol} & \textbf{Name} & \textbf{Description} \\ 
    \midrule

    \(\bm{z}^\ell\) 
    & Activity / primal state
    & State of layer \(\ell\). The input and output activities are clamped to
    \(\bm{x}\) and \(\bm{y}\), respectively, while hidden activities are inferred. \\

    \(\bm{u}^\ell\)
    & Dual state
    & Dual coordinate of the activity,
    \(\bm{u}^\ell=\phi^{-1}(\bm{z}^\ell)\), equivalently
    \(\bm{z}^\ell=\phi(\bm{u}^\ell)\). \\

    \(\bm{a}^\ell\)
    & Preactivation
    & Bottom-up linear input to layer \(\ell\),
    \(\bm{a}^\ell=\bm{W}^\ell\bm{z}^{\ell-1}\).
    At feedforward initialisation, \(\bm{a}^\ell=\bm{u}^\ell\). \\

    \(\widehat{\bm{z}}^\ell\)
    & Predicted activity
    & Bottom-up prediction of the activity at layer \(\ell\),
    \(\widehat{\bm{z}}^\ell=\phi(\bm{a}^\ell)\). \\

    \(\phi\)
    & Activation function
    & Map \(\phi:\mathbb{R}^n\rightarrow\phi(\mathbb{R}^n)\) applied elementwise. \\

    \(\psi\) 
    & Scalar Bregman potential
    & Strictly convex scalar function matched to the activation through
    \(\psi'(z)=\phi^{-1}(z)\). \\

    \(\Psi\)
    & Layerwise Bregman potential
    & Separable extension of \(\psi\) to vector-valued activities,
    \(\Psi(\bm{z})=\sum_i\psi(z_i)\). \\

    \(D_\Psi(\bm{p},\bm{q})\)
    & Bregman divergence
    & Non-negative discrepancy generated by \(\Psi\) between
    \(\bm{p}\) and \(\bm{q}\). \\

    \bottomrule
    \end{tabularx}
\end{table}

\subsection{Predictive coding networks}
\label{sec:pcns}
While PCNs can be defined on arbitrary graph topologies \citep{salvatori2022learning}, here we consider feedforward networks or multi-layer perceptrons (MLPs) with \(L\) layers. We denote the state or activity of layer \(\ell\) by \(\bm{z}^\ell\) and the weights connecting layer \(\ell\) to layer \(\ell+1\) by \(\bm{W}^\ell\). Without loss of generality, we omit biases and consider a single data sample. For supervised learning, the input and output activities are clamped to some data so that \(\bm{z}^0 \leftarrow \bm{x}\) and \(\bm{z}^L \leftarrow \bm{y}\), while the hidden states \(\{\bm{z}^\ell\}_{\ell=1}^{L-1}\) are free to evolve during inference, as explained below.

\paragraph{Energy function.} We write the bottom-up prediction of the activity of layer \(\ell\) as
\begin{align}
    \widehat{\bm{z}}^\ell =
    \phi (\bm{a}^\ell), \qquad \bm{a}^\ell = \bm{W}^\ell\bm{z}^{\ell-1},
    \label{eq:pc-prediction}
\end{align}
where \(\bm{a}^\ell\) is the corresponding preactivation, and \(\phi(\cdot)\) is an elementwise activation function such as \(\tanh\). Standard PC defines the layerwise prediction error as\footnote{In some works, the activation function is applied before the weighted sum, \(\bm{W}^\ell \phi\left(\bm{z}^{\ell-1} \right)\). This removes the activation derivative from the weight, but not the activity, update.}
\begin{align}
    \bm{\varepsilon}^\ell \coloneqq \bm{z}^\ell - \widehat{\bm{z}}^\ell = \bm{z}^\ell - \phi \left( \bm{W}^\ell \bm{z}^{\ell-1} \right),
    \label{eq:pc-error}
\end{align}
and minimises the sum of squared prediction errors
\begin{align}
    \mathcal{F}(\bm{z},\bm{\theta}) = \frac{1}{2} \sum_{\ell=1}^{L} \left\| \bm{\varepsilon}^\ell \right\|_2^2, 
    \label{eq:pc-energy}
\end{align}
where we collect all the states in \(\bm{z} \coloneqq \{\bm{z}^\ell\}_{\ell=0}^{L}\), and the weights in \(\bm{\theta} \coloneqq \{\bm{W}^\ell\}_{\ell=1}^{L}\). Eq.~\ref{eq:pc-energy} is an energy function in two separate senses: (i) it minimises the so-called variational free energy (equal to the negative evidence lower bound) under certain assumptions \citep{buckley2017free, bogacz2017tutorial}; and (ii) both the inference and learning dynamics of PC can be derived as gradient flows on the energy, as we review below. Note that prediction errors are measured using the \textit{squared Euclidean distance}.

\paragraph{Inference dynamics.} Before updating the weights, the free, hidden activities \(\ell\in\{1,\dots,L-1\}\) are inferred via gradient descent (GD) on the energy (Eq.~\ref{eq:pc-energy}). In continuous time, the inference dynamics are then given by
\begin{align}
    \frac{d\bm{z}^\ell}{d\tau} = - \nabla_{\bm{z}^\ell} \mathcal{F} = - \bm{\varepsilon}^\ell + \textcolor{orange!75!black}{\phi'\left(\bm{a}^{\ell+1}\right)} \odot \left(\bm{W}^{\ell+1}\right)^\top \bm{\varepsilon}^{\ell+1},
    \label{eq:pc-inference}
\end{align}
where \(\tau\) represents inference time, \(\textcolor{orange!75!black}{\phi'(\cdot)}\) is the elementwise derivative of the activation function, and \(\odot\) denotes elementwise multiplication. We highlight the activation derivative since it is the object that we aim to avoid explicitly computing. The inference dynamics are local in the sense that updating a hidden layer requires information only from that layer and its immediate neighbours. We typically initialise the activities with a feedforward pass and run the dynamics until an approximate equilibrium 
\begin{align}
    \bm{z}^{\star} \approx \arg\min_{\bm{z}} \mathcal{F}(\bm{z},\bm{\theta}),
    \label{eq:pc-approx-equilib}
\end{align}
which should be numerically close to the unique global minimum for linear networks \citep{innocenti2025mu}, with \(\phi=\bm{I}\), or any critical point for nonlinear networks.

\paragraph{Learning dynamics.} At the converged values of the inferred activities (Eq.~\ref{eq:pc-approx-equilib}), the weights are updated by stochastic GD on the same energy (Eq.~\ref{eq:pc-approx-equilib}). Again in continuous time, the learning dynamics are
\begin{align}
    \frac{d\bm{W}^\ell}{dt} = -\eta\,\nabla_{\bm{W}^\ell}\mathcal{F}(\bm{z}^\star) = \eta \, \textcolor{orange!75!black}{\phi'\left(\bm{a}^\ell\right)} \odot
    \bm{\varepsilon}^\ell
    \left(\bm{z}^{\ell-1}\right)^\top,
    \label{eq:pc-learning}
\end{align}
where \(\eta\) is the learning rate. Note that we do not backpropagate through the inner, inference dynamics (Eq.~\ref{eq:pc-approx-equilib}). \S\ref{sec:pc-implicit-grad} explains why this choice is justified when an (approximate) equilibrium is reached. Similar to the states, the weight updates are local in that they require only the presynaptic activity and quantities available at the postsynaptic layer.

\paragraph{On analog implementation.} The standard formulation of PC reviewed above highlights a potential difficulty for physical implementations of nonlinear PCNs. In particular, although a nonlinear activation function \(\phi\) may arise naturally from the input-output response of an analog device, the inference and learning dynamics of Eqs.~\ref{eq:pc-inference} \& \ref{eq:pc-learning} always explicitly require its derivative \(\textcolor{orange!75!black}{\phi'}\). In \S\ref{sec:bregman-pc}, we show that this dependence can be removed by replacing the squared Euclidean prediction errors with Bregman divergences matched to the particular activation function.

\subsection{Bregman divergences}
\label{sec:bregman-divergences}
\paragraph{Definition.} A Bregman divergence generalises the squared Euclidean distance by using the geometry of a convex function \citep{bregman1967relaxation}. Using standard notation, let \(\Psi:\mathcal{D}\rightarrow\mathbb{R}\) be a differentiable and strictly convex function on a convex domain \(\mathcal{D}\). The Bregman divergence generated
by \(\Psi\) is
\begin{align}
    D_\Psi(\bm{p},\bm{q}) &\coloneqq \Psi(\bm{p}) - \Psi(\bm{q}) - \nabla_{\bm{q}}\Psi(\bm{q})^\top (\bm{p}-\bm{q}).
\label{eq:bregman-divergence}
\end{align}
The last two terms in Eq.~\ref{eq:bregman-divergence} are the first-order Taylor approximation of \(\Psi(\bm{p})\) around \(\bm{q}\). The Bregman divergence \(D_\Psi(\bm{p},\bm{q})\) therefore measures the gap between the value of the convex function at \(\bm{p}\) and the tangent-plane approximation at \(\bm{q}\) (see Figure~\ref{fig:tanh-bregman-divergence}c for an example). 

\paragraph{Properties.} Since the function \(\Psi\) is strictly convex by assumption, the divergence cannot be negative \(D_\Psi(\bm{p},\bm{q})\geq0\), and is exactly zero when \(\bm{p}=\bm{q}\). \(\Psi\) is also referred to as the Bregman potential, since it is the function from which the divergence is derived (for key notation, see Table~\ref{tab:key-notation}). Finally, note that Bregman divergences are not, in general, symmetric and therefore need not define a metric.

\paragraph{Example: squared Euclidean distance.} Consider the scalar quadratic potential \(\psi(p)=\frac{1}{2}p^2\). Since \(\psi'(q)=q\), substituting into the Bregman divergence's definition (Eq.~\ref{eq:bregman-divergence}) gives
\begin{align}
    D_\psi(p,q) = \frac{1}{2}p^2 - \frac{1}{2}q^2 - q(p-q) = \frac{1}{2}(p-q)^2.
    \label{eq:bregman-quadratic-example}
\end{align}
For the squared vector norm \(\Psi(\bm{p})=\frac{1}{2}\|\bm{p}\|_2^2\), this immediately generalises to
\begin{align}
    D_\Psi(\bm{p},\bm{q}) = \frac{1}{2} \|\bm{p}-\bm{q}\|_2^2.
    \label{eq:bregman-euclidean}
\end{align}
Thus, the quadratic prediction errors of the standard PC energy (Eq.~\ref{eq:pc-energy}) can be seen as special Bregman divergences with a squared Euclidean norm as generating or potential function. For other examples of Bregman divergences (e.g. the Kullback–Leibler divergence), see \S\ref{sec:other-bregman-examples}.

\section{Predictive Coding with Bregman Divergences}
\label{sec:bregman-pc}
\begin{figure}[t]
    \vskip 0.2in
    \begin{center}
        \centerline{\includegraphics[width=\textwidth]{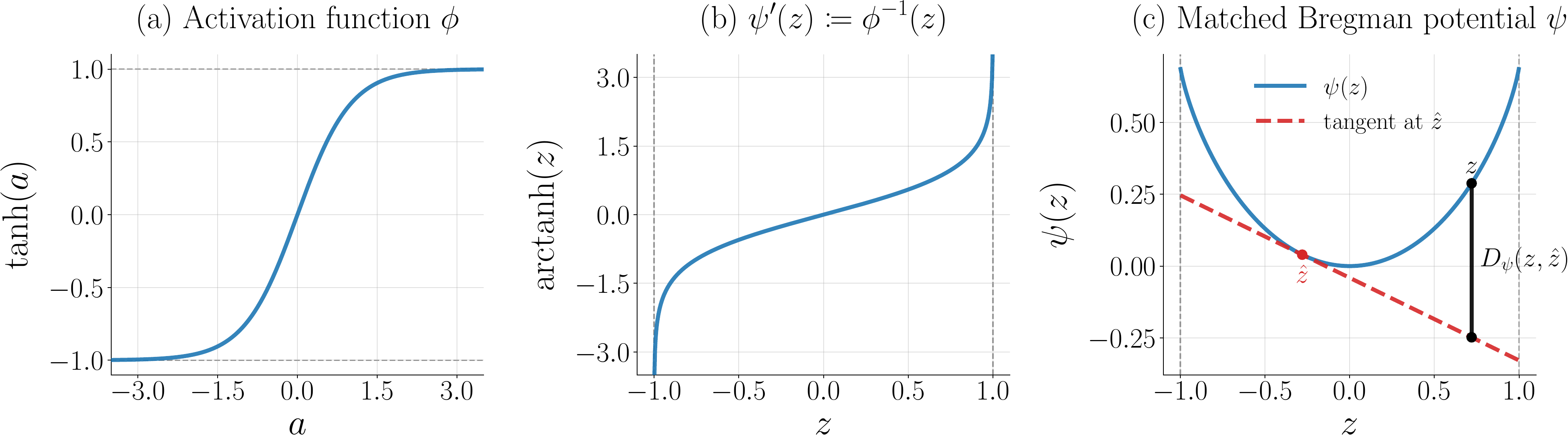}}
        \caption{\textbf{Constructing an activation-matched Bregman divergence for \(\tanh\).} \textbf{(a)} The activation function \(\phi(a)=\tanh(a)\). \textbf{(b)} Its inverse, \(\phi^{-1}(z)=\operatorname{arctanh}(z)\), which defines the derivative of the matching potential through \(\psi'(z)=\phi^{-1}(z)\) (Eq.~\ref{eq:matching-scalar-potential}). \textbf{(c)} Integrating the inverse activation gives the matching potential \(\psi(z)=z\operatorname{arctanh}(z)+\frac{1}{2}\log(1-z^2)\), up to an additive constant. Since \(\operatorname{arctanh}(z)\) is strictly increasing, \(\psi\) is strictly convex. The Bregman divergence \(D_\psi(z,\widehat z)\) is the gap between \(\psi(z)\) and the tangent-line approximation of \(\psi\) at the prediction \(\widehat z=\phi(a)\). See Figures~\ref{fig:sigmoid-bregman-divergence}-\ref{fig:linear-bregman-divergence} for divergences with other activations.}
        \label{fig:tanh-bregman-divergence}
    \end{center}
    \vskip -0.25in
\end{figure}
In this section, we develop a generalisation of PC in which nonlinear prediction errors are measured using Bregman divergences matched to the activation function, instead of squared Euclidean norms. Closely related activation-matched losses have previously been used to avoid explicit activation derivatives in parameter updates \citep{wang2020generalised, amid2022locoprop, wang2023lifted}. ``LocoProp'' uses these losses to optimise layerwise weights towards targets obtained from backpropagation \citep{amid2022locoprop}, while ``lifted Bregman training'' introduces hidden activities as auxiliary (PC-like) optimisation variables in a closely related layerwise Bregman objective \citep{wang2023lifted}. Here, we instead formulate these variables as PC states and derive local mirror-descent inference dynamics before parameter learning. Throughout, we use \(\tanh\) as an example activation (Figure~\ref{fig:tanh-bregman-divergence}a) and provide results for other suitable activations in the Appendix.


\paragraph{Activation function assumptions.} We assume that \(\phi\) is differentiable with \(\phi'(a)>0\), so that it is strictly increasing and invertible on its range, and the matched potential defined below (Eq.~\ref{eq:matching-scalar-potential}) is strictly convex. These assumptions include, for example, \(\tanh\), sigmoid, and softplus activations. Non-smooth or non-monotonic activations require a generalised treatment, which we discuss in \S\ref{sec:discussion}.

\paragraph{Activation-matched Bregman potential.} We define the matched Bregman scalar potential \(\psi:\phi(\mathbb{R})\rightarrow\mathbb{R}\) by requiring \textit{its derivative to equal the inverse of the activation function} (see Figure~\ref{fig:tanh-bregman-divergence}a-b for an example),
\begin{align}
    \psi'(z) \coloneqq \phi^{-1}(z).
    \label{eq:matching-scalar-potential}
\end{align}
Equivalently, \(\psi\) is any antiderivative of \(\phi^{-1}\), since additive constants do not affect the resulting Bregman divergence. Because \(\phi\) is strictly increasing by assumption, its inverse \(\phi^{-1}\) is also strictly increasing. Therefore, \(\psi'\) is strictly increasing and \(\psi\) is strictly convex (Figure~\ref{fig:tanh-bregman-divergence}c), which is the key property needed to define a Bregman divergence, as reviewed in \S\ref{sec:bregman-divergences}.

\begin{wrapfigure}{r}{0.48\columnwidth}
    \vskip 0.25in
    \centering
    \includegraphics[width=\linewidth]{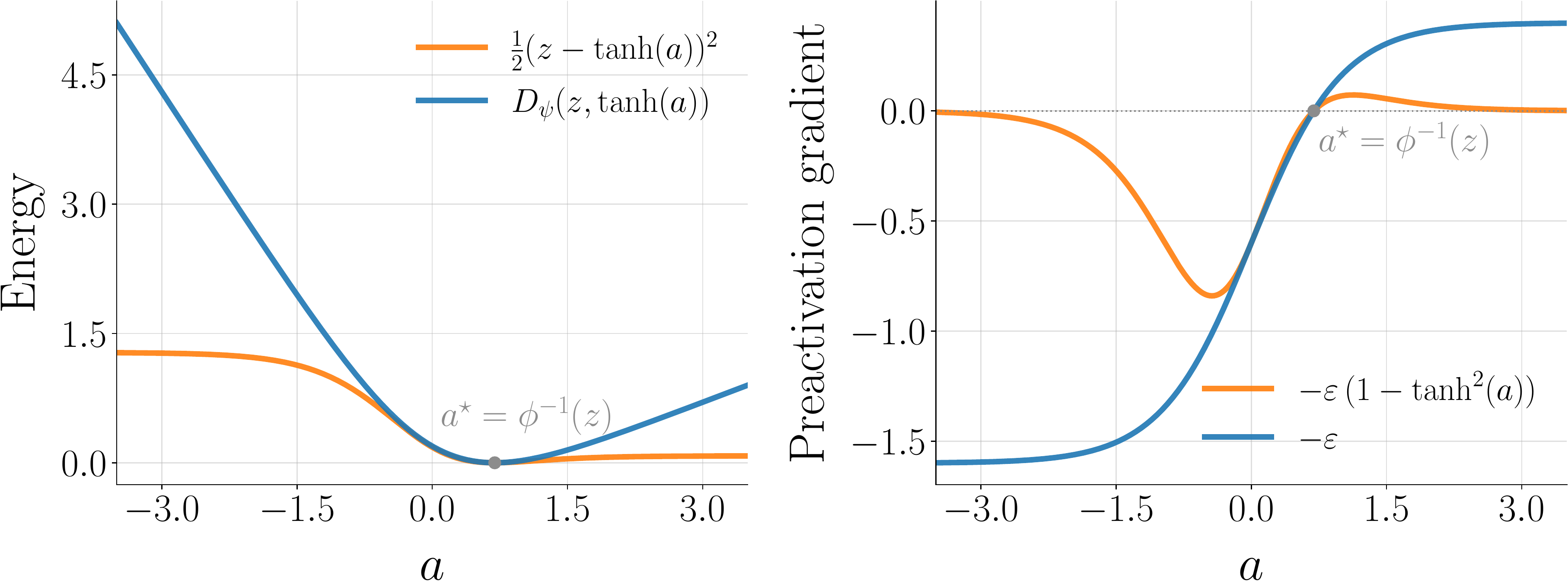}
    \caption{\textbf{Activation matching prevents attenuation of prediction errors.} For \(\tanh\), the standard squared error gives \(\partial_a\mathcal{F} = - \varepsilon\phi'(a)\), which is attenuated near saturation. The matched Bregman divergence contributes \(1/\phi'(a)\), cancelling this factor and yielding \(\partial_a D_\psi = -\varepsilon\). Analogous results for other activations are shown in Figures~\ref{fig:sigmoid-preact-deriv}-\ref{fig:linear-preact-deriv}.}
    \label{fig:tanh-preact-deriv}
    \vskip 0.1in
\end{wrapfigure}

\paragraph{Why matching removes the activation derivative.} To see the effect of the activation-matching condition of Eq.~\ref{eq:matching-scalar-potential}, consider a scalar prediction \(\widehat z=\phi(a)\) with prediction error \(\varepsilon \coloneqq z-\phi(a)\). As clear from the standard PC inference and learning dynamics (Eqs.~\ref{eq:pc-inference} \& \ref{eq:pc-learning}), differentiating the
squared prediction error with respect to the preactivation introduces the activation derivative:
\begin{align}
    \frac{\partial}{\partial a} \left(z-\phi(a)\right)^2/2 = -\varepsilon\,\phi'(a).
    \label{eq:standard-preact-derivative}
\end{align}
Thus, the prediction-error signal is modulated by \(\phi'(a)\), and can be strongly attenuated when the activation saturates
(see Figure~\ref{fig:tanh-preact-deriv}). For the activation-matched Bregman divergence \(D_\psi(z,\phi(a))\), instead,
\begin{align}
    \frac{\partial}{\partial a}D_\psi\left(z,\phi(a)\right) = -\psi''\left(\phi(a)\right)\varepsilon\,\phi'(a).
\end{align}
By the matching condition \(\psi'=\phi^{-1}\) (Eq.~\ref{eq:matching-scalar-potential}) and the inverse-function theorem,
\begin{align}
    \psi''\left(\phi(a)\right) = (\phi^{-1})'\left(\phi(a)\right) = \frac{1}{\phi'(a)}.
\end{align}
Hence, the factor \(\phi'(a)\) introduced by differentiating the nonlinear prediction is exactly cancelled by the reciprocal factor arising from the curvature of the matched potential:
\begin{align}
    \frac{\partial}{\partial a}D_\psi\left(z,\phi(a)\right) = -\varepsilon.
    \label{eq:bregman-preact-derivative}
\end{align}
Activation matching therefore encodes the nonlinearity into the geometry of the error measure itself, so that the preactivation derivative depends directly on the error without requiring explicit evaluation of \(\phi'(a)\).

\paragraph{Vector-valued activities.} For multidimensional activities \(\bm{z}\in\mathbb{R}^n\), we apply the same construction elementwise and define the separable potential
\begin{align}
    \Psi: \phi(\mathbb{R})^n\rightarrow\mathbb{R}, \quad \Psi(\bm{z}) \coloneqq \sum_i \psi(z_i).
\end{align}
Its gradient is therefore the elementwise inverse activation,
\begin{align}
    \nabla_{\bm{z}} \Psi(\bm{z}) = 
    \begin{pmatrix}
        \phi^{-1}(z_1) \\
        \vdots \\
        \phi^{-1}(z_n)
    \end{pmatrix} = \phi^{-1}(\bm{z}).
    \label{eq:matching-vector-potential}
\end{align}
Hence, the corresponding activation-matched divergence satisfies \(\nabla_{\bm{a}}D_\Psi\left(\bm{z},\phi(\bm{a})\right) = -\bm{\varepsilon}\).

\paragraph{Bregman-PC energy.} We now use the activation-matched divergence to construct a nonlinear PC energy whose inference and learning dynamics do not explicitly contain \(\phi'\). In particular, we replace the layerwise squared prediction error of standard PC (Eq.~\ref{eq:pc-energy}) with
\begin{align}
    \mathcal{F}^\ell_{\mathrm{B}} \coloneqq D_{\Psi} \left( \bm{z}^\ell, \phi(\bm{a}^\ell) \right), \quad \nabla_{\bm{z}} \Psi(\bm{z}) =\phi^{-1}(\bm{z}).
    \label{eq:bregman-layer-energy}
\end{align}
For a network with a conventional linear readout \(\bm{a}^L = \bm{W}^L \bm{z}^{L-1}\), we allow an arbitrary output loss \(\mathcal{L}(\bm{a}^L,\bm{y})\) and define the total Bregman-PC energy as
\begin{empheq}[box=\highlightbox]{align}
    \mathcal{F}_{\mathrm{B}} = \sum_{\ell=1}^{L-1} D_{\Psi} \left( \bm{z}^\ell, \phi(\bm{a}^\ell) \right) + \mathcal{L} \left( \bm{a}^L,\bm{y} \right).
    \label{eq:bregman-energy}
\end{empheq}
Each hidden term is non-negative and exactly zero at the feedforward pass, as in standard PC. This formulation accommodates, for example, a linear readout trained with either mean-squared error (MSE) or softmax cross-entropy.\footnote{A Bregman divergence can also be used at the output whenever the output activation and target domain admit an appropriate matching potential.} For convenience, we define the output error as the negative output-loss gradient, \(\bm{\varepsilon}^L \coloneqq - \nabla_{\bm{a}^L}\mathcal{L}\), so that for example, \(\bm{\varepsilon}^L = \bm{y}-\bm{a}^L\) for MSE. 

The Bregman-PC energy (Eq.~\ref{eq:bregman-energy}) itself can be algebraically more involved than the quadratic energy of standard PC (Eq.~\ref{eq:pc-energy}), since the matched potential is obtained by integrating the inverse activation, \(\psi(z)=\int^z \phi^{-1}(s)ds\). For \(\tanh\), for example, \(\psi(z) = z \operatorname{arctanh}(z) +\frac{1}{2}\log(1-z^2)\), up to an additive constant. Crucially, however, neither this antiderivative nor the inverse activation needs to be explicitly evaluated in the local inference and learning dynamics derived below. Note that our formulation also remains compatible with equilibrium propagation \citep{scellier2017equilibrium}, by introducing a nudging parameter \(\beta\) scaling the output loss \(\mathcal{L}\).

\paragraph{Primal and dual neural states.} The activation-matched Bregman potential \(\Psi\) (Eq.~\ref{eq:matching-vector-potential}) also induces a natural change of coordinates for the neural activities. We refer to \(\bm{z}^\ell\) as the \textit{primal} state and define its corresponding \textit{dual} state as
\begin{align}
    \bm{u}^\ell &\coloneqq \nabla_{\bm{z}}\Psi \left(\bm{z}^\ell\right)
    = \phi^{-1} \left(\bm{z}^\ell\right), \qquad \bm{z}^\ell = \phi\left(\bm{u}^\ell\right)
    \label{eq:dual-state}
\end{align}
Thus, \(\bm{u}^\ell\) and \(\bm{z}^\ell\) describe the same neural state in two coordinate systems related by the geometry of \(\Psi\). Under forward-pass initialisation, \(\bm{z}^\ell=\phi(\bm{a}^\ell)\), and therefore
\(\bm{u}^\ell=\bm{a}^\ell\). In other words, the dual state is equal to the feedforward preactivation at initialisation.

\begin{wrapfigure}{r}{0.48\columnwidth}
    \centering
    \includegraphics[width=\linewidth]{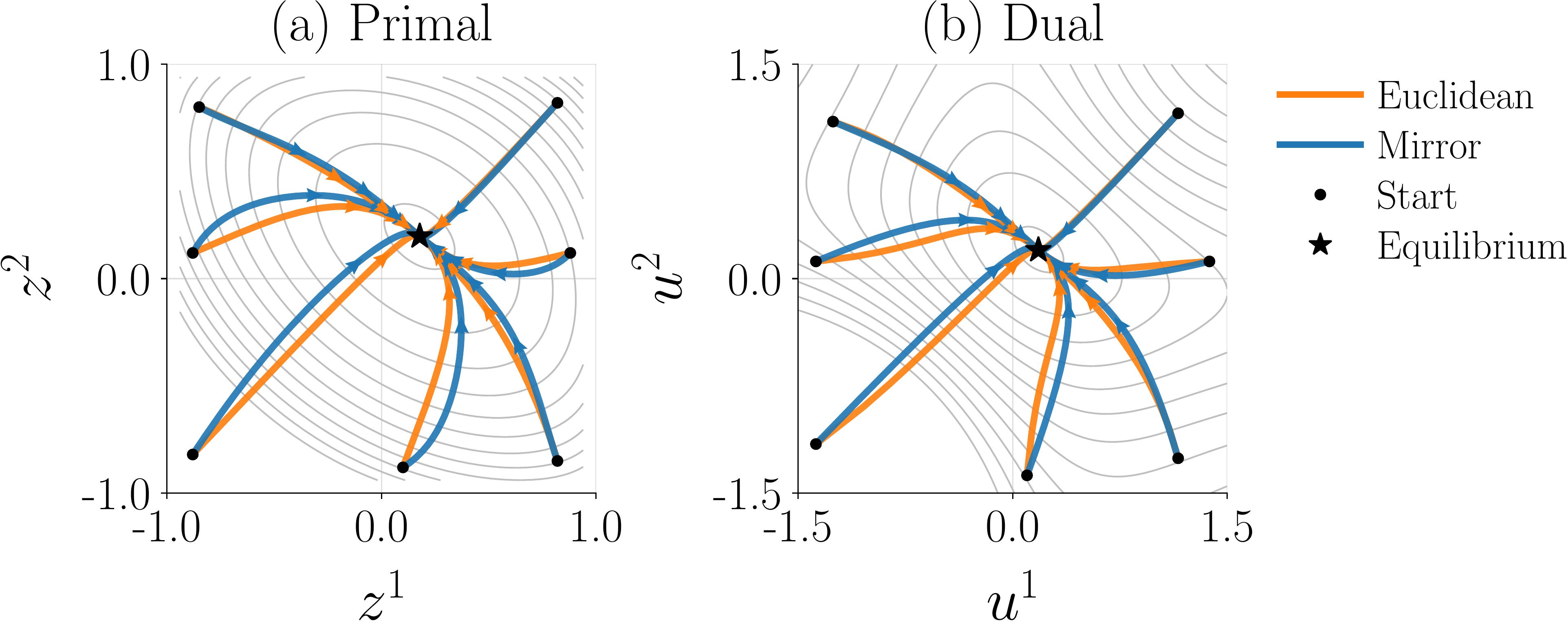}
    \caption{\textbf{Mirror inference in primal and dual coordinates.} Inference trajectories for a two-hidden-state Bregman PCN with \(\phi=\tanh\), initialised from the same state and, in this example, converging to the same equilibrium. \textbf{(a)} In primal coordinates \(\bm{z}=(z^1,z^2)\), Euclidean inference follows \(d\bm{z}/d\tau=-\nabla_{\bm{z}}\mathcal{F}_{\mathrm{B}}\), whereas mirror inference follows the preconditioned flow \(d\bm{z}/d\tau = -[\nabla^2\Psi(\bm{z})]^{-1}\nabla_{\bm{z}}\mathcal{F}_{\mathrm{B}}\), leading to a different trajectory through activity space. \textbf{(b)} In dual coordinates \(\bm{u}=\nabla\Psi(\bm{z})=\phi^{-1}(\bm{z})\), the same mirror dynamics take the simpler form \(d\bm{u}/d\tau=-\nabla_{\bm{z}}\mathcal{F}_{\mathrm{B}}\). Grey contours show the Bregman-PC energy in the corresponding coordinates. For other activations, see Figures~\ref{fig:sigmoid-flows}-\ref{fig:linear-flows}.}
    \label{fig:tanh-flows}
    \vskip -0.5in
\end{wrapfigure}

\paragraph{Mirror-descent inference.} We now use these dual coordinates to formulate inference as \textit{mirror descent}. In particular, rather than taking an ordinary gradient step directly in the primal activity \(\bm{z}^\ell\), we evaluate the energy gradient with respect to \(\bm{z}^\ell\) but evolves its dual coordinate \(\bm{u}^\ell=\nabla_{\bm{z}}\Psi(\bm{z}^\ell)\). In continuous time,
\begin{align}
    \frac{d\bm{u}^\ell}{d\tau} = \frac{d}{d\tau} \nabla_{\bm{z}}\Psi(\bm{z}^\ell) \coloneqq -\nabla_{\bm{z}^\ell}\mathcal{F}_{\mathrm{B}},
    \label{eq:mirror-grad-flow}
\end{align}
where recall that \(\tau\) denotes inference time. Since \(\bm{z}^\ell = \phi(\bm{u}^\ell)\), the same dynamics can be expressed in primal coordinates as
\begin{align}
    \frac{d\bm{z}^\ell}{d\tau} = -\left[\nabla_{\bm{z}}^2\Psi(\bm{z}^\ell)\right]^{-1} \nabla_{\bm{z}^\ell}\mathcal{F}_{\mathrm{B}} = - \operatorname{diag}\left(\phi'(u_i^\ell)\right) \nabla_{\bm{z}^\ell}\mathcal{F}_{\mathrm{B}},
    \label{eq:bregman-primal-flow}
\end{align}
where the second equality follows from the activation-matching relation \(\bm{z}=\phi(\bm{u})\). Mirror inference is therefore equivalent to a \textit{preconditioned gradient flow} in the primal activities, where the preconditioner is the inverse local curvature of the matched potential \(\Psi\). Thus, mirror flow changes the trajectory and local speed of inference relative to ordinary Euclidean gradient flow, as illustrated in Figure~\ref{fig:tanh-flows}.

For the Bregman-PC energy (Eq.~\ref{eq:bregman-energy}), the dual-coordinate dynamics (Eq.~\ref{eq:mirror-grad-flow}) take the particularly simple form:
\begin{empheq}[box=\highlightbox]{align}
    \frac{d\bm{u}^\ell}{d\tau} = - \bm{u}^\ell + \bm{a}^\ell + \left(\bm{W}^{\ell+1}\right)^\top \bm{\varepsilon}^{\ell+1}.
    \label{eq:bregman-inference}
\end{empheq}
See Appendix~\ref{sec:state-updates-deriv} for the derivation. The update depends only on the current dual state, its linear bottom-up prediction, and a weighted prediction error from the next layer. Importantly, an analog implementation can evolve \(\bm{u}^\ell\) directly and obtain the primal activity from the activation, \(\bm{z}^\ell=\phi(\bm{u}^\ell)\). Inference therefore requires neither an explicit activation derivative \(\phi'\) nor an explicit evaluation of the inverse activation \(\phi^{-1}\). A candidate analog circuit implementing these dynamics is described in Appendix~\ref{sec:bregman-pc-circuit}.

\paragraph{Energy descent.} Although mirror descent follows a different trajectory from Euclidean gradient flow (Figure~\ref{fig:tanh-flows}), it still monotonically decreases the Bregman-PC energy. Since \(\Psi\) is convex, its inverse Hessian is positive semidefinite wherever it is defined \(\left[\nabla_{\bm{z}}^2\Psi(\bm{z}^\ell)\right]^{-1}\succeq0\), and hence
\begin{align}
    \frac{d\mathcal{F}_{\mathrm{B}}}{d\tau} &= \sum_{\ell=1}^{L-1} \left( \nabla_{\bm{z}^\ell}\mathcal{F}_{\mathrm{B}} \right)^\top \frac{d\bm{z}^\ell}{d\tau} = -\sum_{\ell=1}^{L-1} \left( \nabla_{\bm{z}^\ell}\mathcal{F}_{\mathrm{B}} \right)^\top \left[\nabla_{\bm{z}}^2\Psi(\bm{z}^\ell)\right]^{-1} \left( \nabla_{\bm{z}^\ell}\mathcal{F}_{\mathrm{B}} \right) \leq 0.
    \label{eq:bregman-energy-descent}
\end{align}
Thus, while mirror inference changes the path and speed of inference, it preserves descent of the same Bregman-PC energy.

\paragraph{Weight updates.} The matched PC-Bregman energy (Eq.~\ref{eq:bregman-energy}) also removes the activation derivative from the learning dynamics (derivation in \S\ref{sec:weight-updates-deriv}):
\begin{empheq}[box=\highlightbox]{align}
    \frac{d \bm{W}^\ell}{dt} = -\eta \,\nabla_{\bm{W}^\ell} \mathcal{F}_{\mathrm{B}} = \eta \,\bm{\varepsilon}^\ell \left(\bm{z}^{\ell-1}\right)^\top,
    \label{eq:bregman-weight-update}
\end{empheq}
which has the same local outer-product form between the postsynaptic prediction error and the presynaptic activity as standard linear PC. 

\paragraph{Recovery of standard linear PC.} Our proposed Bregman-PC formulation recovers standard linear PC exactly. Specifically, if \(\phi(\bm{a})=\bm{a}\), then \(\phi^{-1}(\bm{a})=\bm{a}\) and the matching potential is the squared norm \(\Psi(\bm{z}) = \frac{1}{2}\|\bm{z}\|_2^2\). As shown in \S\ref{sec:bregman-divergences}, the layerwise matched Bregman divergence then reduces to the standard squared prediction error:
\begin{align}
    D_{\Psi}
    \left( \bm{z}^\ell, \bm{W}^\ell \bm{z}^{\ell-1} \right)
    = \frac{1}{2} \left\| \bm{z}^\ell - \bm{W}^\ell \bm{z}^{\ell-1} \right\|_2^2.
\end{align}
Moreover, the Hessian is the identity, the primal and dual coordinates coincide \(\bm{u}^\ell = \bm{z}^\ell\), and mirror inference reduces to ordinary Euclidean gradient flow (see Figures~\ref{fig:linear-bregman-divergence}, \ref{fig:linear-preact-deriv} \& \ref{fig:linear-flows}). Bregman PC therefore reduces exactly to standard PC in the linear case, while providing an activation-matched nonlinear generalisation in which neither inference nor learning requires explicit derivatives of the activation function. This recovery is useful because most PC theory has been developed for linear networks yet shown to predict and control behaviour of nonlinear ones \citep{ishikawa2024local, innocenti2024only, innocenti2026infinite, seely2026augmented}, as we further demonstrate experimentally in the next section.

\section{Experiments}
\label{sec:exps}
\begin{table}[t]
    \centering
    \caption{\textbf{Classification and generation performance of Bregman PC vs BP and standard PC.} Validation classification accuracy and image-reconstruction MSE on MNIST and Fashion-MNIST (mean \(\pm\) SEM over 3 seeds). See Figures~\ref{fig:classification-metrics}-\ref{fig:generation-examples} for loss curves and examples of generated images, and \S\ref{exp-details} for experimental details.}
    \label{tab:benchmark-performance}
    \small
    \begin{tabular}{lcccc}
    \toprule
     & \multicolumn{2}{c}{Classification accuracy (\%) \(\uparrow\)}
     & \multicolumn{2}{c}{Reconstruction MSE \([0,1]\) \(\downarrow\)} \\
    \cmidrule(lr){2-3}\cmidrule(lr){4-5}
    Algorithm & MNIST & Fashion-MNIST & MNIST & Fashion-MNIST \\
    \midrule
    BP
    & \(98.05 \pm 0.10\)
    & \(88.92 \pm 0.11\)
    & \(0.05310 \pm 0.00003\)
    & \(0.05263 \pm 0.00004\) \\
    Standard PC
    & \(98.08 \pm 0.08\)
    & \(88.87 \pm 0.10\)
    & \(0.05320 \pm 0.00003\)
    & \(0.05275 \pm 0.00006\) \\
    Bregman PC
    & \(97.57 \pm 0.08\)
    & \(87.30 \pm 0.24\)
    & \(0.05334 \pm 0.00007\)
    & \(0.05326 \pm 0.00006\) \\
    \bottomrule
    \end{tabular}
\end{table}
We now show that Bregman PC remains practically competitive with standard PC while still preserving its key computational properties. We focus on two main questions: (i) whether Bregman PC can train nonlinear networks with performance comparable to standard PC; and (ii) whether it retains previously established learning regimes for PC.

\paragraph{Performance on simple tasks.} We first compare Bregman PC against standard PC and BP on standard classification and generative learning tasks using MNIST and Fashion-MNIST, following the experimental setting of \cite{pinchetti2024benchmarking}. Across these tasks, Bregman PC reaches performance comparable to both standard PC and BP (see Table~\ref{tab:benchmark-performance}), indicating that replacing squared prediction errors with activation-matched Bregman divergences, and performing inference in the dual coordinates, does not significantly hurt performance.

\begin{wrapfigure}{r}{0.55\columnwidth}
    \centering
    \includegraphics[width=\linewidth]{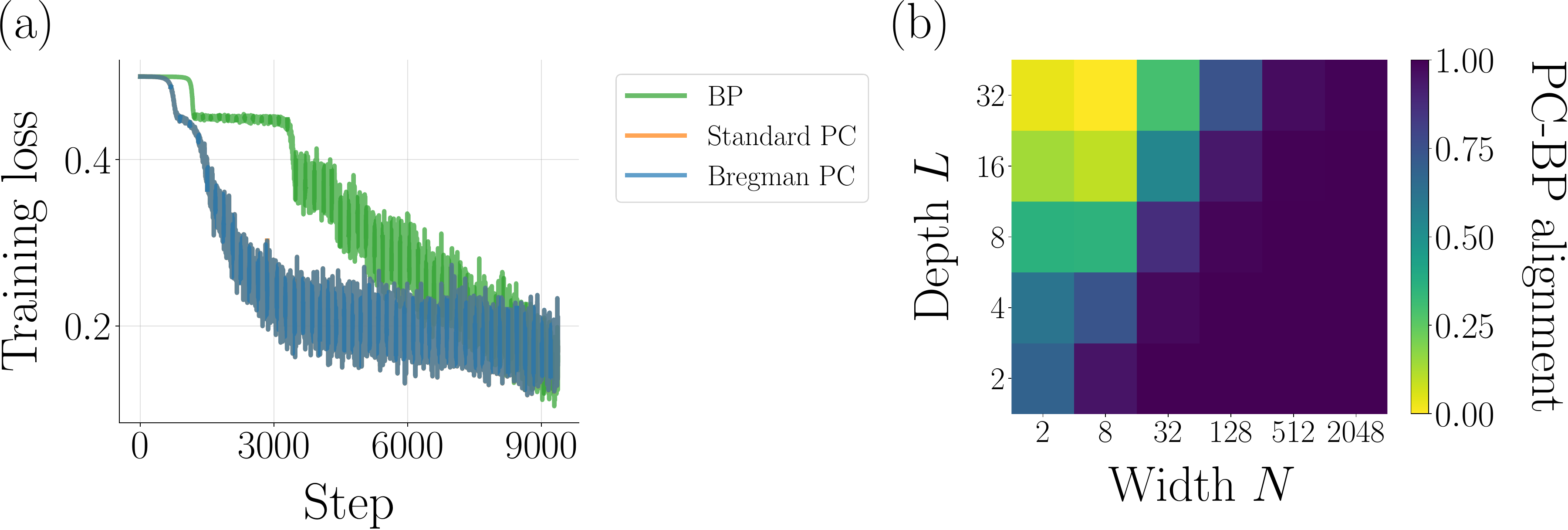}
    \caption{\textbf{Bregman PC preserves established PC learning regimes.}
    \textbf{(a)} ``Saddle-to-saddle dynamics'' for 5-layer \(\tanh\) MLPs trained on MNIST with a small, near-origin initialisation. BP remains on a prolonged high-loss plateau, whereas standard PC and Bregman PC rapidly escape, with nearly overlapping loss trajectories. \textbf{(b)} Cosine similarity between Bregman-PC and BP parameter gradients after \(100\) training steps for nonlinear residual networks trained on CIFAR-10 using the width- and depth-stable parameterisation of \cite{innocenti2026infinite}. Similarities are averaged over three random seeds and approach \(1\) as width becomes large relative to depth. For more details, see \S\ref{exp-details}.}
    \label{fig:pc-regimes}
    \vskip -0.25in
\end{wrapfigure}

\paragraph{Preservation of established PC learning regimes.} We next ask whether Bregman PC preserves some learning regimes previously established for standard PC, ranging from dynamics that differ qualitatively from BP to a regime in which PC approaches BP. First, we reproduced an experiment from \cite{innocenti2024only}, training nonlinear MLPs on MNIST from a small, near-origin initialisation. In this regime, BP remains trapped for an extended period near the highly degenerate saddle at the origin, whereas standard PC rapidly escapes, showing fast ``saddle-to-saddle'' dynamics. Bregman PC reproduces this behaviour, escaping on a similar timescale and following a nearly indistinguishable loss trajectory from standard PC (Figure~\ref{fig:pc-regimes}a).
At the other extreme, we test whether Bregman PC also preserves a recently established regime where the PC gradients converge to those computed by BP under stable large-model parameterisations \citep{ishikawa2024local, innocenti2026infinite}. Following \cite{innocenti2026infinite}, we trained nonlinear residual networks using a width- and depth-stable parameterisation on CIFAR-10, varying both network width \(N\) and depth \(L\). Consistent with previous results, we find that the Bregman-PC gradients align with the BP gradients as the model width becomes much larger than the depth (Figure~\ref{fig:pc-regimes}b). 

Together, these results show that Bregman PC preserves both the distinctive learning dynamics of standard PC, in the form of fast saddle-to-saddle dynamics, while recovering its asymptotic convergence to BP for large, stably parameterised models.

\section{Discussion}
\label{sec:discussion}
\paragraph{Summary.} We introduced an activation-matched Bregman formulation of PC in which inference and learning avoid explicit evaluation of activation-function derivatives (Figure~\ref{fig:standard-vs-bregman}). By choosing the Bregman potential such that \(\nabla\Psi=\phi^{-1}\), the derivative of the nonlinearity is absorbed into the geometry of the prediction error. Combined with mirror-descent inference in the corresponding dual coordinates, this yields local derivative-free dynamics while recovering standard PC exactly in the linear case. Our experiments show that this modification preserves both competitive learning performance and key computational regimes of standard PC.

\subsection{Limitations and future directions}
\paragraph{Hardware implementation and validation.} Our main motivation is analog implementation. Our formulation should therefore be understood as making nonlinear PC \textit{more hardware-compatible}, rather than providing a complete analog implementation. The inference and learning dynamics of Bregman PC require only weighted interactions, local prediction errors, state integration, and evaluation of the activation function itself. Appendix~\ref{sec:bregman-pc-circuit} illustrates how the inference dynamics could be mapped onto a simple analog circuit. Whether the proposed formulation is sufficient for an effective analog implementation of PC remains to be established and should ultimately be validated in hardware, including under device noise and mismatch. We also note that analog friendliness is related to, but distinct from, biological plausibility: while both often favour local and distributed computation, operations that are convenient to implement physically need not be biologically realistic, and vice versa.

\paragraph{Activation-function and empirical limitations.} A limitation of our construction is the assumption that the activation is differentiable and strictly increasing. This includes common saturating nonlinearities such as \(\tanh\) and sigmoid, but excludes standard ReLU and globally non-monotonic activations such as GELU and SiLU. More general Bregman constructions based on proximal maps can accommodate non-smooth and non-invertible monotone activations such as ReLU \citep{wang2023lifted}. Extending the mirror-descent PC dynamics developed here to this setting is therefore a natural direction for future work, while globally non-monotonic activations will require a different treatment. Our empirical evaluation is also limited to relatively small feedforward networks and standard benchmark tasks, and testing whether the same behaviour persists in larger and more complex architectures will be important.


\subsection*{AI use statement}
In this work, we used generative AI tools mainly for writing code to implement all the experiments and for editing the paper to improve clarity. 
Specifically, we used AI tools for formulating certain mathematical claims (the descent result of Eq.~\ref{eq:bregman-energy-descent}), designing and providing feedback on experiments, and implementing methods.
We have not used generative AI tools for developing theoretical models or conceptual frameworks, providing critical ingredients for proving mathematical claims, proposing or refining hypotheses. Creating synthetic data sets, assisting in the writing of proofs or translation, cleaning or reformatting datasets, and supporting qualitative and thematic data analysis are not applicable to this work.
Additionally, we used generative AI tools for suggesting experimental hyperparameters, brainstorming and editing. We have reviewed all AI-assisted work. In particular, LLM-generated code was verified and tested for correctness. We take responsibility for the final content of this work,
including text, claims or artifacts produced with the aid of generative AI.

\subsection*{Reproducibility statement}
Experimental details are provided in \S\ref{exp-details}, and code to reproduce all the results is available at \url{https://github.com/thebuckleylab/jpc/tree/main/experiments/bregman_pc}.

\subsubsection*{Acknowledgments}
This work was supported by funding from the Wellcome Trust grant 313955/Z/24/Z. I would like to thank Marc Gong Bacvanski, Bryce Primavera, Benjamin Scellier, and Elene Lominadze for discussions that motivated this work. Some of the simulations were supported by Google Cloud Credits awarded by the Google TPU Research Cloud Programme (2026).

\bibliography{iclr2027_conference}
\bibliographystyle{iclr2027_conference}

\newpage
\appendix
\section*{Appendix} \label{appendix}
\startcontents[appendix]
\printcontents[appendix]{}{1}{}
\bigskip

\section{On predictive coding and ``implicit gradients''} 
\label{sec:pc-implicit-grad}
In our review of PCNs in \S\ref{sec:pcns}, we pointed out that the weight update of PC (Eq.~\ref{eq:pc-learning}) does not backpropagate through the inference updates of Eq.~\ref{eq:pc-inference}, effectively treating the activities as constant. This can be justified if (i) the two levels of optimisation minimise the same objective (as in PC), and (ii) we converge to a critical point of the inner problem, since 
\begin{align}
    \frac{\partial \mathcal{F}(\bm{z}^\star(\bm{\theta}), \bm{\theta})}{\partial \bm{\theta}}
    &= \underbrace{\vphantom{\cancelto{0}{\frac{\partial \mathcal{F}}{\partial \bm{h}^\star(\bm{\theta})}}}\frac{\partial \mathcal{F}}{\partial \bm{\theta}}}_{\text{explicit}} + \underbrace{\cancelto{0}{\frac{\partial \mathcal{F}}{\partial \bm{z}^\star(\bm{\theta})}}\frac{\partial\bm{z}^\star(\bm{\theta})}{\partial \bm{\theta}}}_{\text{implicit}},
    \label{eq:total-pc-weight-deriv}
\end{align}
where we define \(\bm{z}^\star(\bm{\theta}) = \argmin_{\bm{z}} \mathcal{F}(\bm{z}, \bm{\theta})\). We emphasise that the activity solution is a function of the weights by writing \(\bm{z}(\bm{\theta})\). Eq.~\ref{eq:total-pc-weight-deriv} shows the total weight derivative of the energy at a critical point of the activities, where \(\partial \mathcal{F}/\partial \bm{z}^\star = \bm{0}\). By the product rule, this decomposes into two terms: (i) the direct or \textit{explicit} dependence of the energy on the weights (treating the activities as constant), and (ii) the indirect or \textit{implicit} effect of the weights on the energy through the activity equilibrium.

Because (i) inference and learning minimise the same objective (i.e. the energy), and because (ii) we assume convergence on the inner problem, the implicit gradient vanishes (Eq. \ref{eq:total-pc-weight-deriv}). We therefore do not need to backpropagate through the PC inference updates and can just take the direct (local) weight gradient. This contrasts with related models such as deep equilibrium models \citep{bai2019deep}, neural differential equations \citep{kidger2022neural} and modern Hopfield networks \citep{hoover2023energy}, which use either backpropagation through time or other techniques to estimate the implicit gradient. We also note that this property is incompatible with equilibrium propagation \citep{scellier2017equilibrium}, which introduces a cost or loss function in addition to an energy.

\section{Other examples of Bregman divergences} 
\label{sec:other-bregman-examples}
\paragraph{Kullback-Leibler (KL) divergence.} Consider two discrete probability distributions \(\bm{p}\) and \(\bm{q}\), and let the generating function be the negative (Shannon) entropy
\begin{align}
    \Psi(\bm{p}) = \sum_i p_i \log p_i.
\end{align}
Its gradient has components \([\nabla\Psi(\bm{q})]_i = 1+\log q_i\). The corresponding Bregman divergence (Eq.~\ref{eq:bregman-divergence}) then is
\begin{align}
    D_\Psi(\bm{p},\bm{q}) &= \sum_i p_i\log p_i - \sum_i q_i\log q_i \notag \\
    &\qquad - \sum_i(1+\log q_i)(p_i-q_i) \\
    &= \sum_i p_i\log\frac{p_i}{q_i} = D_{\mathrm{KL}}(\bm{p}\|\bm{q}),
    \label{eq:bregman-kl}
\end{align}
where in the second equality we used \(\sum_i p_i=\sum_i q_i=1\). Thus, the KL divergence between probability distributions is a Bregman divergence generated by the negative entropy.

\paragraph{Bernoulli KL divergence.} For a scalar \(p\in(0,1)\), consider the negative binary entropy
\begin{align}
    \psi(p) = p\log p + (1-p)\log(1-p).
\end{align}
Notably, its derivative is the inverse sigmoid or logit function
\begin{align}
    \psi'(q) = \log\frac{q}{1-q} = \sigma^{-1}(q) = \operatorname{logit}(q),
\end{align}
where \(\sigma(p) = 1/(1+e^{-p})\). This is the key property we use in \S\ref{sec:bregman-pc} for constructing Bregman divergences whose geometry is matched to an activation function. The corresponding Bregman divergence is
\begin{align}
    D_\psi(p,q) &= \psi(p)-\psi(q)-\psi'(q)(p-q) \\
    &= p\log\frac{p}{q} + (1-p)\log\frac{1-p}{1-q} \\
    &= D_{\mathrm{KL}} \bigl(\operatorname{Bern}(p) \,\|\, \operatorname{Bern}(q) \bigr).
    \label{eq:bregman-bernoulli-kl}
\end{align}
The Bernoulli KL divergence is therefore generated by the negative binary entropy.

\section{Derivation of Bregman-PC updates} 
\label{sec:updates-deriv}

\subsection{State updates} 
\label{sec:state-updates-deriv}
Here we derive the Bregman-PC inference dynamics of Eq.~\ref{eq:bregman-inference}. Recall from Eq.~\ref{eq:bregman-energy} that the Bregman-PC energy is
\begin{align}
    \mathcal{F}_{\mathrm{B}}
    = \sum_{\ell=1}^{L-1} D_{\Psi}\left( \bm{z}^\ell, \phi(\bm{a}^\ell) \right) + \mathcal{L}(\bm{a}^L,\bm{y}).
\end{align}
For a hidden layer \(\ell\), the activity \(\bm{z}^\ell\) appears in two terms of the energy: (i) its local prediction-error term \(\mathcal{F}_{\mathrm{B}}^\ell\), and (ii) the prediction-error term of the subsequent layer \(\mathcal{F}_{\mathrm{B}}^{\ell+1}\). Hence, the activity gradient is given by
\begin{align}
    \nabla_{\bm{z}^\ell}\mathcal{F}_{\mathrm{B}} = \nabla_{\bm{z}^\ell}\mathcal{F}_{\mathrm{B}}^\ell + \nabla_{\bm{z}^\ell}\mathcal{F}_{\mathrm{B}}^{\ell+1}.
\end{align}
\paragraph{Local contribution.} From the definition of a Bregman divergence (Eq.~\ref{eq:bregman-divergence}), its gradient with respect to the first argument is
\begin{align}
    \nabla_{\bm{p}}D_\Psi(\bm{p},\bm{q}) = \nabla\Psi(\bm{p})-\nabla\Psi(\bm{q}).
\end{align}
Setting \(\bm{p}=\bm{z}^\ell\) and \(\bm{q}=\phi(\bm{a}^\ell)\) therefore gives
\begin{align}
    \nabla_{\bm{z}^\ell}\mathcal{F}_{\mathrm{B}}^\ell =
    \nabla\Psi(\bm{z}^\ell) - \nabla\Psi\left(\phi(\bm{a}^\ell)\right) = \phi^{-1}(\bm{z}^\ell) - \phi^{-1}\left(\phi(\bm{a}^\ell)\right) = \bm{u}^\ell-\bm{a}^\ell,
    \label{eq:bregman-local-state-gradient}
\end{align}
where we used \(\bm{u}^\ell=\nabla\Psi(\bm{z}^\ell)=\phi^{-1}(\bm{z}^\ell)\).
\paragraph{Feedback contribution.} The activity \(\bm{z}^\ell\) enters the subsequent layer only through the next preactivation \(\bm{a}^{\ell+1} = \bm{W}^{\ell+1}\bm{z}^\ell\). Applying the chain rule therefore gives
\begin{align}
    \nabla_{\bm{z}^\ell}\mathcal{F}_{\mathrm{B}}^{\ell+1}
    &=
    \left(
    \frac{\partial\bm{a}^{\ell+1}}
    {\partial\bm{z}^\ell}
    \right)^\top
    \nabla_{\bm{a}^{\ell+1}}
    \mathcal{F}_{\mathrm{B}}^{\ell+1} =
    \left(\bm{W}^{\ell+1}\right)^\top
    \nabla_{\bm{a}^{\ell+1}}
    \mathcal{F}_{\mathrm{B}}^{\ell+1}.
\end{align}
Using the activation-matching property of Eq.~\ref{eq:bregman-preact-derivative} derived in \S\ref{sec:bregman-pc}, we get
\begin{align}
    \nabla_{\bm{z}^\ell}\mathcal{F}_{\mathrm{B}}^{\ell+1} = - \left(\bm{W}^{\ell+1}\right)^\top \bm{\varepsilon}^{\ell+1}.
    \label{eq:bregman-feedback-state-gradient}
\end{align}
For the final hidden layer, the same expression follows from our definition \(\bm{\varepsilon}^{L}\coloneqq
-\nabla_{\bm{a}^{L}}\mathcal{L}\). Substituting the two gradient contributions from Eqs.~\ref{eq:bregman-local-state-gradient} and \ref{eq:bregman-feedback-state-gradient} into the mirror-gradient-flow dynamics \(d\bm{u}^\ell/d\tau \coloneqq -\nabla_{\bm{z}^\ell}\mathcal{F}_{\mathrm{B}}\) gives
\begin{align}
    \boxed{
        \frac{d\bm{u}^\ell}{d\tau} = - \bm{u}^\ell+\bm{a}^\ell + \left(\bm{W}^{\ell+1}\right)^\top \bm{\varepsilon}^{\ell+1}
    },
\end{align}
which is Eq.~\ref{eq:bregman-inference}.

\subsection{Weight updates} 
\label{sec:weight-updates-deriv}
We next derive the Bregman-PC weight updates of Eq.~\ref{eq:bregman-weight-update}. For a hidden layer \(\ell\in\{1,\dots,L-1\}\), the weights \(\bm{W}^\ell\) enter the energy only through the corresponding preactivation \(\bm{a}^\ell = \bm{W}^\ell \bm{z}^{\ell-1}\). Therefore, applying the chain rule gives
\begin{align}
    \nabla_{\bm{W}^\ell}\mathcal{F}_{\mathrm{B}}
    = \nabla_{\bm{a}^\ell}\mathcal{F}_{\mathrm{B}}^\ell
    \left(\bm{z}^{\ell-1}\right)^\top.
\end{align}
Using the activation-matching property of Eq.~\ref{eq:bregman-preact-derivative}, \(\nabla_{\bm{a}^\ell}\mathcal{F}_{\mathrm{B}}^\ell = -\bm{\varepsilon}^\ell\), we obtain
\begin{align}
    \boxed{
    \frac{d\bm{W}^\ell}{dt} = -\eta\nabla_{\bm{W}^\ell}\mathcal{F}_{\mathrm{B}} = \eta \,\bm{\varepsilon}^\ell \left(\bm{z}^{\ell-1}\right)^\top
    },
\end{align}
which is Eq.~\ref{eq:bregman-weight-update}. The same update holds for the output-layer weights as \(\bm{\varepsilon}^L \coloneqq -\nabla_{\bm{a}^L}\mathcal{L}\).

\section{Candidate analog implementation of Bregman-PC inference}
\label{sec:bregman-pc-circuit}
\begin{figure}[t]
    \centering
    \begin{circuitikz}[
        scale=0.90,
        transform shape,
        block/.style={
            draw,
            rounded corners,
            minimum width=1.4cm,
            minimum height=0.65cm,
            align=center,
            font=\small
        },
        >=Latex
    ]
    
    
    \node (a) at (0,0) {$a_i^\ell$};
    
    \node[circ] (unode) at (3.0,0) {};
    \node[above=1.5pt of unode, xshift=-10pt] {$u_i^\ell$};
    
    \draw
        (a)
        to[R,l=$R$]
        (unode);
    
    \draw
        (unode)
        to[C,l_=$C$]
        (3.0,-1.8)
        node[ground] {};
    
    
    \node[block] (phiu) at (3.0,1.6)
        {$\phi$};
    
    \node (z) at (3.0,2.8)
        {$z_i^\ell$};
    
    \draw[-Latex]
        (unode) -- (phiu);
    
    \draw[-Latex]
        (phiu) -- (z);
    
    \node[block] (phia) at (0,1.6)
        {$\phi$};
    
    \node (zhat) at (0,2.8)
        {$\widehat z_i^\ell$};
    
    \draw[-Latex]
        (a) -- (phia);
    
    \draw[-Latex]
        (phia) -- (zhat);
    
    
    \node (eps) at (6.0,2.0)
        {$\bm{\varepsilon}^{\ell+1}$};
    
    \node[block] (gm) at (6.0,0)
        {
            $(\bm{W}^{\ell+1})^\top$\\
            weighted sum
        };
    
    \draw[-Latex]
        (eps) -- (gm.north);
    
    \draw[-Latex]
        (gm.west)
        -- node[above,font=\scriptsize]
           {$I_{\mathrm{fb},i}^\ell$}
        (unode);
    
    \node[
        align=center,
        font=\scriptsize
    ] at (6.0,-1.0)
    {
        no activation derivative\\
        required
    };
    
    
    \node[
        align=center,
        font=\small
    ] at (3.6,-3.0)
    {
    $\displaystyle
    I_{\mathrm{fb},i}^\ell
    =
    \frac{1}{R}
    \sum_j
    W_{ji}^{\ell+1}
    \varepsilon_j^{\ell+1}
    $
    };
    
    \node[
        align=center,
        font=\small
    ] at (3.6,-4.0)
    {
    $\displaystyle
    RC\,\frac{d u_i^\ell}{dt}
    =
    -u_i^\ell+a_i^\ell
    +
    \sum_j
    W_{ji}^{\ell+1}
    \varepsilon_j^{\ell+1}
    $
    };
    \end{circuitikz}
    \caption{\textbf{Candidate analog implementation of Bregman-PC inference.}
    The capacitor voltage represents the dual state \(u_i^\ell\). A resistor connecting it to the bottom-up preactivation \(a_i^\ell\) generates the local restoring current \((a_i^\ell-u_i^\ell)/R\), while a weighted-sum circuit supplies a feedback current proportional to \(\sum_j W_{ji}^{\ell+1}\varepsilon_j^{\ell+1}\). The same nonlinear transfer function can be used to obtain the activity \(z_i^\ell=\phi(u_i^\ell)\) and prediction \(\widehat z_i^\ell=\phi(a_i^\ell)\). Kirchhoff's current law reproduces the Bregman-PC inference dynamics in Eq.~\ref{eq:bregman-inference} up to an overall rescaling of time by \(RC\). Unlike standard nonlinear PC (Figure~\ref{fig:standard-vs-bregman}), the feedback pathway requires no explicit evaluation of the activation derivative.}
    \label{fig:bregman-pc-circuit}
    \vskip -0.25in
\end{figure}
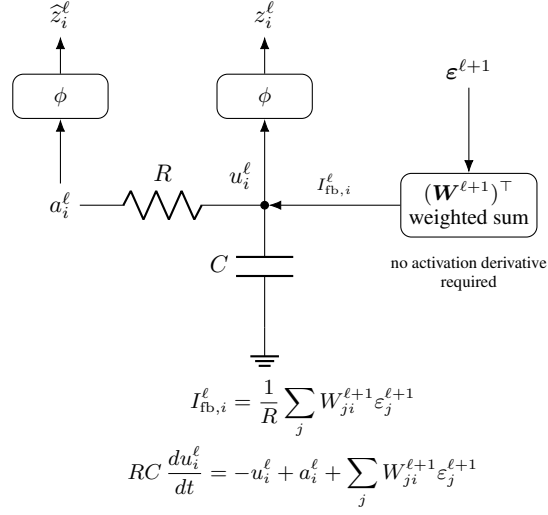
The derivative-free dynamics of Bregman PC (Eq.~\ref{eq:bregman-inference}) admit a simple analog interpretation. Consider the inference dynamics of a single unit \(i\) in layer \(\ell\),
\begin{align}
    \frac{d u_i^\ell}{d\tau} = - u_i^\ell + a_i^\ell +\sum_j W_{ji}^{\ell+1} \varepsilon_j^{\ell+1}.
\end{align}
A candidate circuit implementation is shown in Figure~\ref{fig:bregman-pc-circuit}. The dual state \(u_i^\ell\) is represented by the voltage across a capacitor \(C\), while the bottom-up preactivation \(a_i^\ell\) is represented by a second voltage. Connecting these nodes through a resistor \(R\) generates the current
\begin{align}
    I_{\mathrm{local},i}^\ell = \frac{a_i^\ell-u_i^\ell}{R}.
\end{align}
The weighted feedback term can be supplied as an additional current
\begin{align}
    I_{\mathrm{fb},i}^\ell = \frac{1}{R} \sum_j W_{ji}^{\ell+1}\varepsilon_j^{\ell+1}.
\end{align}
Kirchhoff's current law then gives
\begin{align} 
    C\frac{d u_i^\ell}{dt} &= \frac{a_i^\ell-u_i^\ell}{R} + I_{\mathrm{fb},i}^\ell, \quad
    RC\frac{d u_i^\ell}{dt} = - u_i^\ell+a_i^\ell +\sum_jW_{ji}^{\ell+1}\varepsilon_j^{\ell+1},
\end{align}
where \(t\) denotes physical circuit time. This reproduces Eq.~\ref{eq:bregman-inference} up to an overall rescaling of time by the circuit time constant \(RC\).

The same nonlinear transfer function \(\phi\) can be applied to the state and preactivation voltages to produce the activity \(z_i^\ell=\phi(u_i^\ell)\) and prediction \(\widehat z_i^\ell=\phi(a_i^\ell)\), from which the local prediction error can be formed. Crucially, the feedback pathway requires only a weighted sum of next-layer prediction errors. Standard nonlinear PC would additionally require each error to be multiplied by the activation derivative \(\phi'(\bm a^{\ell+1})\) before this weighted sum is formed.

We emphasise that Figure~\ref{fig:bregman-pc-circuit} is intended only as a schematic circuit interpretation of the Bregman-PC inference dynamics, not as a complete hardware design including the learning dynamics. The circuitry generating the bottom-up weighted sums, local prediction errors, and programmable synaptic weights is also left unspecified and may be realised using different analog technologies.

\section{Experimental details} 
\label{exp-details}
Unless otherwise stated, trained networks used no biases, \(\tanh\) as hidden activation function, and MSE loss. Comparisons between BP, standard PC, and Bregman PC used identical initial weights. Standard PC used Euclidean GD on the activities (Eq.~\ref{eq:pc-inference}), while Bregman PC used the corresponding mirror-descent dynamics (Eq.~\ref{eq:bregman-inference}), with both initialised from the feedforward activities. The inference step size \(\tau\) and number of steps \(T\) are specified separately for each experiment below.

\begin{figure}[H]
    \begin{center}
        \centerline{\includegraphics[width=\textwidth]{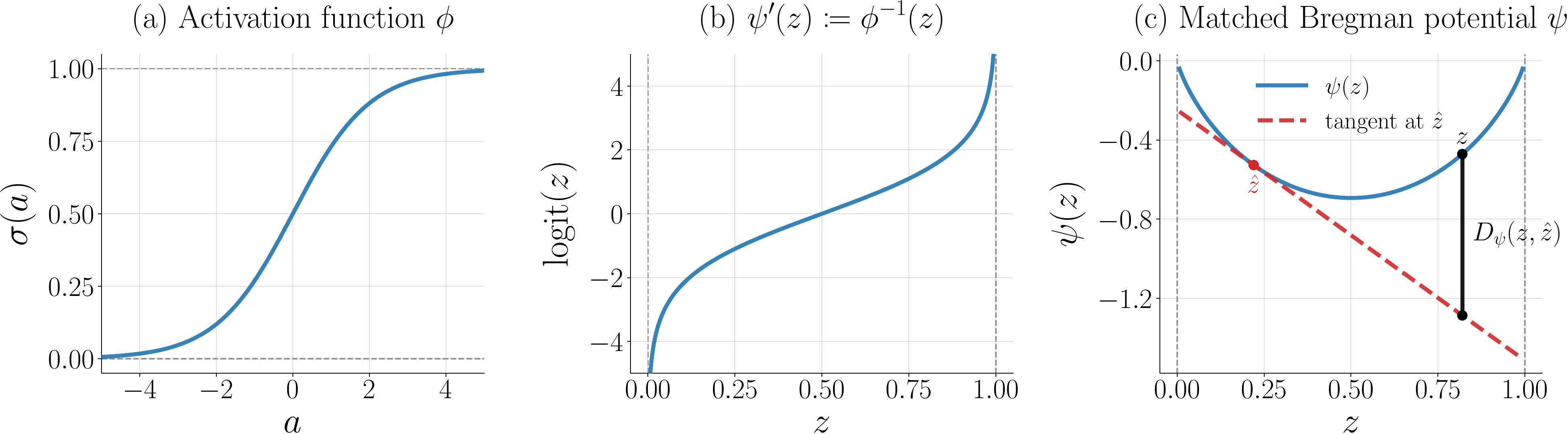}}
        \caption{\textbf{Bregman divergence for the sigmoid function \(\sigmoid\).} \textbf{(a)} The activation function \(\phi(a)=\sigmoid(a)\). \textbf{(b)} Its inverse, \(\sigmoid^{-1}(z)=\operatorname{logit}(z)\). \textbf{(c)} Integrating the inverse activation gives the matching potential \(\psi(z)=z\operatorname{arctanh}(z)+\frac{1}{2}\log(1-z^2)\), up to an additive constant.}
        \label{fig:sigmoid-bregman-divergence}
    \end{center}
\end{figure}

\begin{figure}[H]
    \begin{center}
        \centerline{\includegraphics[width=\textwidth]{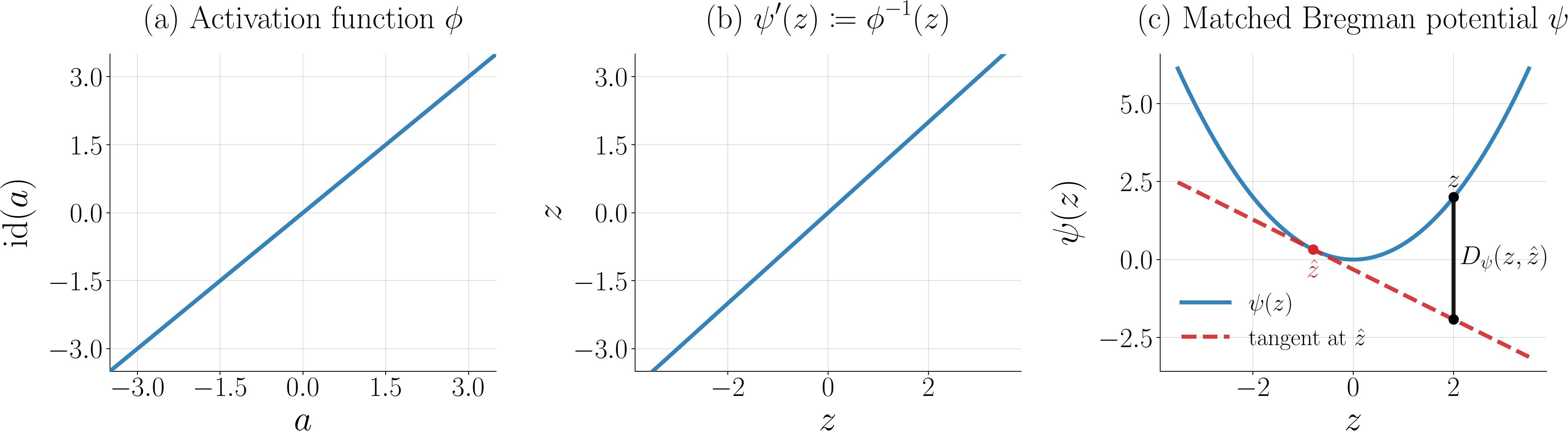}}
        \caption{\textbf{Bregman divergence for the identity function.} \textbf{(a)} The activation function \(\phi(a)=\operatorname{id}(a)\). \textbf{(b)} Its inverse, \(\phi^{-1}(z)=z\). \textbf{(c)} Integrating the inverse activation gives the matching potential \(\psi(z)=z\operatorname{arctanh}(z)+\frac{1}{2}\log(1-z^2)\), up to an additive constant.}
        \label{fig:linear-bregman-divergence}
    \end{center}
\end{figure}

\begin{figure}[H]
    \begin{center}
        \centerline{\includegraphics[width=0.9\textwidth]{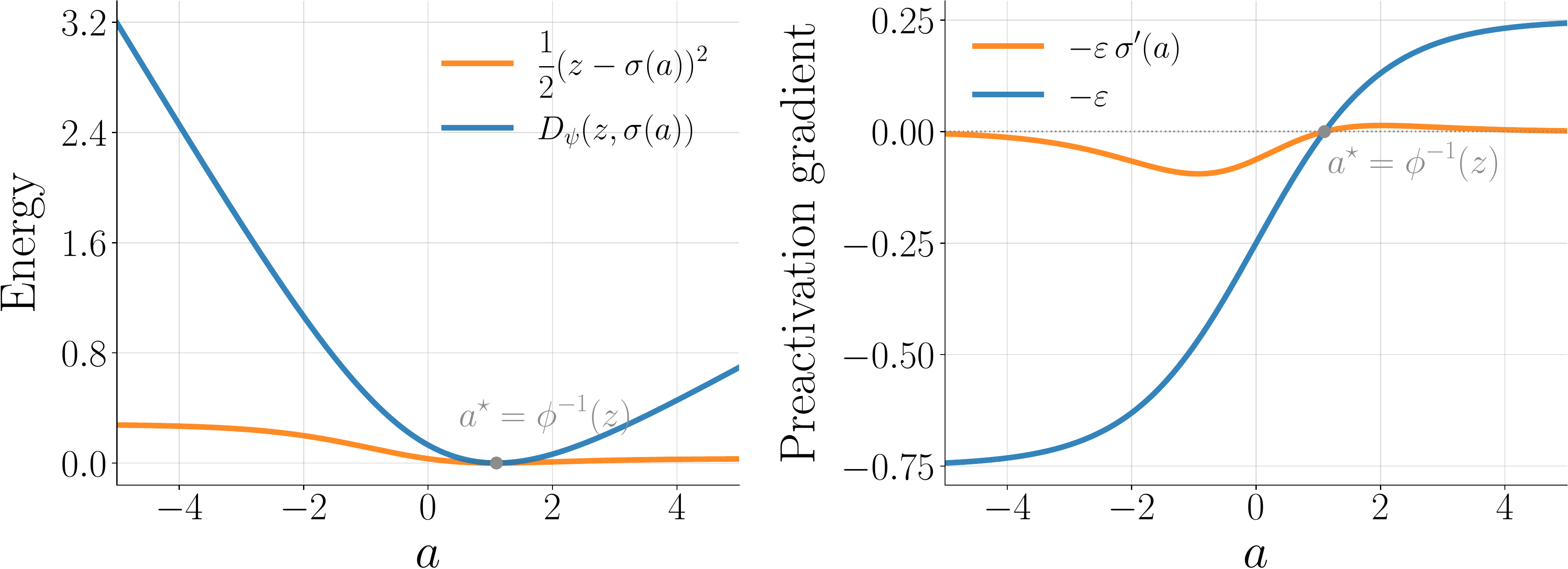}}
        \caption{\textbf{Activation matching for sigmoid.}
        As in Figure~\ref{fig:tanh-preact-deriv}, the standard squared error gives \(\partial_a\mathcal{F} = -\varepsilon\sigma'(a)\), which is attenuated as the activation saturates, whereas the matched Bregman divergence yields \(\partial_aD_\psi = -\varepsilon\).}
        \label{fig:sigmoid-preact-deriv}
    \end{center}
\end{figure}

\begin{figure}[H]
    \begin{center}
        \centerline{\includegraphics[width=0.9\textwidth]{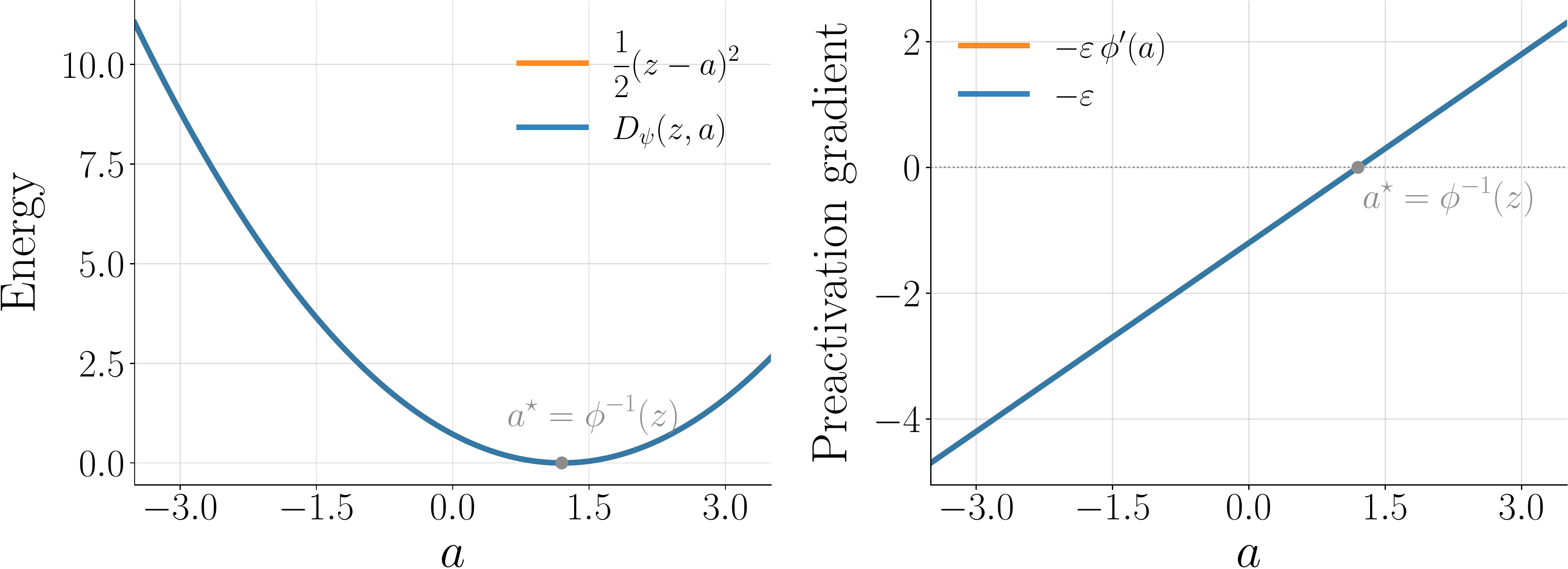}}
        \caption{\textbf{Activation matching for a linear activation.} For \(\phi(a)=a\), we have \(\phi'(a) = 1\), so the standard squared error and
        the matched Bregman divergence coincide, both yielding \(\partial_a\mathcal{F} = \partial_aD_\psi=-\varepsilon\).}
        \label{fig:linear-preact-deriv}
    \end{center}
\end{figure}

\begin{figure}[H]
    \begin{center}
        \centerline{\includegraphics[width=0.9\textwidth]{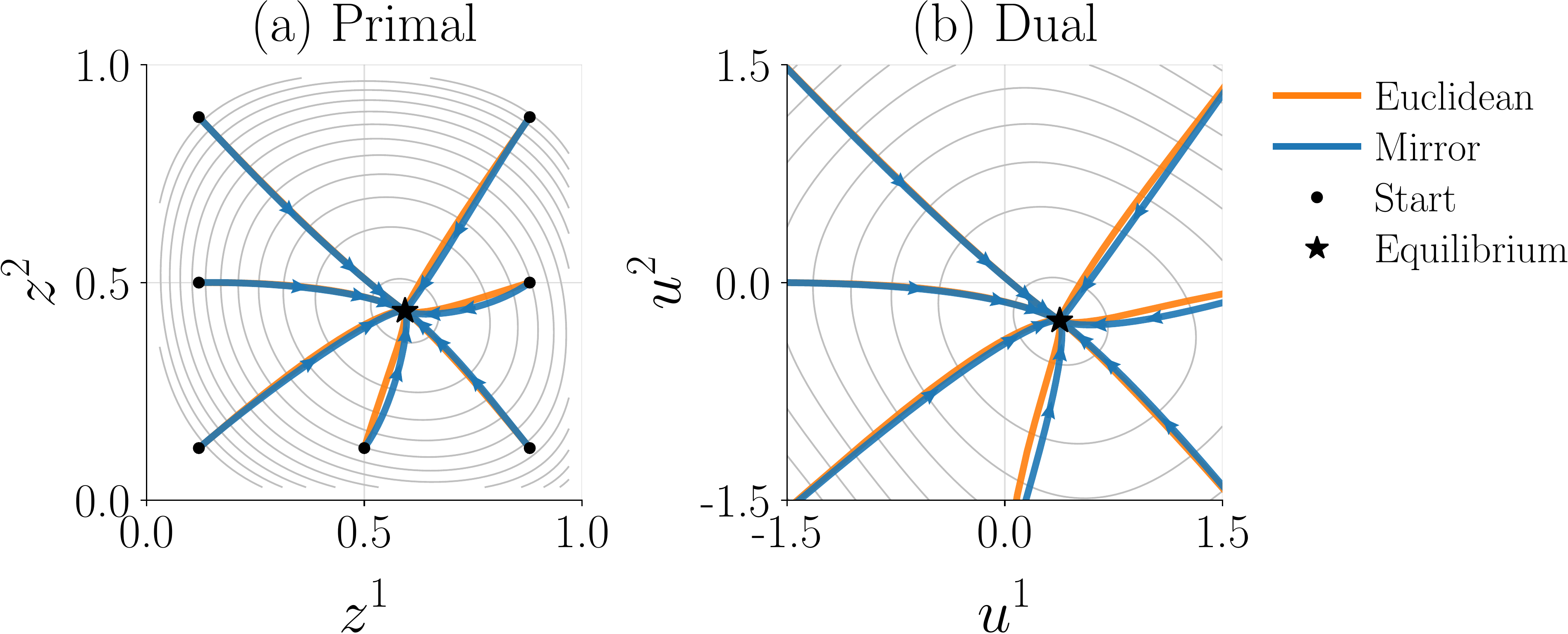}}
        \caption{\textbf{Mirror inference with sigmoid activation.} Analogous to Figure~\ref{fig:tanh-flows} for \(\tanh\), inference trajectories are shown for a two-hidden-state Bregman PCN with a sigmoid activation. Mirror inference follows a curvature-preconditioned gradient flow in primal coordinates and the corresponding simpler dynamics \(d\bm{u}/d\tau=-\nabla_{\bm{z}}\mathcal{F}_{\mathrm B}\) in dual coordinates.}
        \label{fig:sigmoid-flows}
    \end{center}
\end{figure}

\begin{figure}[H]
    \begin{center}
        \centerline{\includegraphics[width=0.9\textwidth]{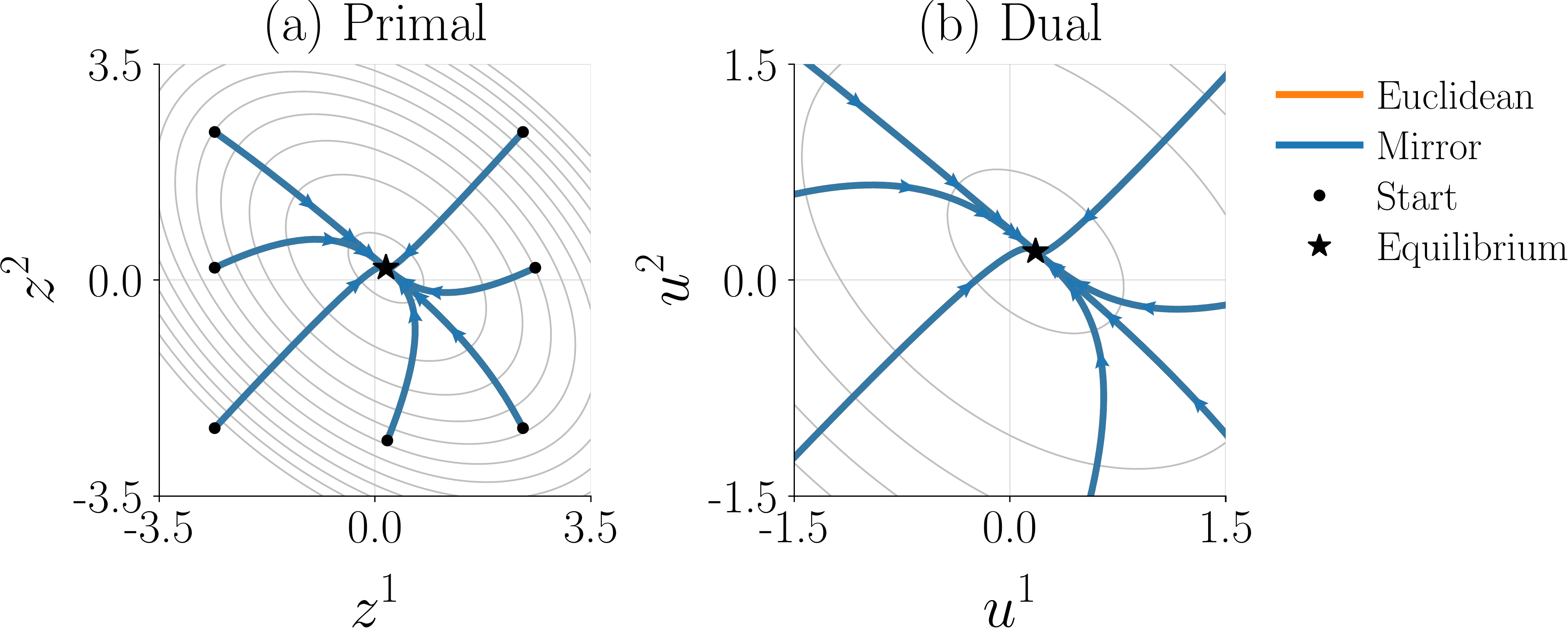}}
        \caption{\textbf{Mirror inference with a linear activation.} Analogous to Figure~\ref{fig:tanh-flows}, but for \(\phi(a)=a\). In this case, the matched Bregman potential is quadratic (Figure~\ref{fig:linear-bregman-divergence}), primal and dual coordinates coincide, and mirror inference reduces exactly to ordinary Euclidean gradient flow, yielding identical trajectories.}
        \label{fig:linear-flows}
    \end{center}
\end{figure}

\paragraph{Toy Bregman geometry and inference trajectories (Figures~\ref{fig:tanh-bregman-divergence}-\ref{fig:tanh-flows} and Figures~\ref{fig:sigmoid-bregman-divergence}--\ref{fig:linear-flows}).} These figures illustrate the activation-matched construction for \(\phi\in\{\tanh,\sigma,\mathrm{id}\}\) and are not trained models. The matched Bregman eneriges (up to an additive constant) are \(\psi(z)=z\operatorname{arctanh}(z)+\frac12\log(1-z^2)\) for \(\tanh\), \(\psi(z)=z\log z+(1-z)\log(1-z)\) for sigmoid, and \(\psi(z)=\frac12 z^2\) for the identity.

Figure~\ref{fig:tanh-flows} compares Euclidean and mirror inference on the same two-hidden-unit Bregman-PC energy
\begin{align}
    \mathcal{F}_{\mathrm{B}}(z^1,z^2) = D_\psi\bigl(z^1,\phi(w_1 x)\bigr) + D_\psi\bigl(z^2,\phi(w_2 z^1)\bigr) + \tfrac12(w_3 z^2-y)^2,
\end{align}
Both flows are integrated with classical RK4 (\(\Delta\tau=0.04\), \(300\) steps) from identical initial activities. Euclidean inference follows \(d\bm{z}/d\tau=-\nabla_{\bm{z}}\mathcal{F}_{\mathrm{B}}\), while mirror inference follows \(d\bm{u}/d\tau=-\nabla_{\bm{z}}\mathcal{F}_{\mathrm{B}}\) with \(\bm{z}=\phi(\bm{u})\). The equilibrium marker is the terminal state of a mirror trajectory, and grey contours are \(12\) equally spaced energy levels between the \(2\)nd and \(92\)nd percentiles of \(\mathcal{F}_{\mathrm{B}}\) on the plotted grid.

\paragraph{Classification and generation tasks (Table~\ref{tab:benchmark-performance}).} We trained fully connected MLPs with two hidden layers of width 256 on MNIST and Fashion-MNIST. Images were flattened and standardised (MNIST: mean \(0.1307\), std.\ \(0.3081\); Fashion-MNIST: mean \(0.5\), std.\ \(0.5\)), while labels were encoded as one-hot vectors. Hyperparameters were selected independently for each algorithm by mean performance on a validation set over three random seeds.

All methods used Adam with no weight decay or learning-rate schedule, batch size \(64\), and \(15\) epochs. We swept the parameter learning rate over \(\eta\in\{5\times 10^{-4},10^{-3},5\times 10^{-3},10^{-2}\}\). For PC inference, we jointly swept the activity step size \(\tau\in\{10^{-3},5\times 10^{-3},10^{-2},5\times 10^{-2},10^{-1},5\times 10^{-1},1\}\) and the number of inference steps \(T\in\{3,5,10,20,50\}\). All methods were evaluated with a single feedforward pass. Table~\ref{tab:benchmark-performance} reports the final classification accuracy and the per-pixel MSE, both averaged over the validation set. For generation, the MSE is in the original \([0,1]\) pixel space so that the two datasets share a common scale.

\paragraph{Saddle-to-saddle dynamics (Figure~\ref{fig:pc-regimes}a).} Following \cite{innocenti2024only}, we trained a \(5\)-layer MLP (\(4\) hidden layers of width \(500\)) on MNIST, comparing BP, standard PC and Bregman PC. Weights were initialised i.i.d. from a zero-mean Gaussian with standard deviation \(\sigma=5\times10^{-3}\). We used SGD, batch size \(64\), learning rate \(\eta=0.05\), and \(35\) epochs. PC inference used \(T=50\) steps with \(\tau=0.1\). Figure~\ref{fig:pc-regimes}a plots the training loss during training.

\paragraph{Bregman-PC convergence to BP (Figure~\ref{fig:pc-regimes}b).} We trained residual MLPs on a single CIFAR-10 batch of size \(64\), using the \(\mu\)PC parameterisation of \cite{innocenti2026infinite}. In particular, weights are initialised from a standard Gaussian, residual branches are scaled by \(1/\sqrt{NL}\), and the network output by \(1/N\). We compared all algorithms over the grid \(N\in\{2,8,32,128,512,2048\}\) and \(L\in\{2,4,8,16,32\}\). All methods used Adam with base learning rate \(10^{-3}\), rescaled as \(\eta/\sqrt{NL}\), for \(100\) full-batch updates. PC inference used \(\tau=0.3\) and \(T=100(L-1)\) steps. The heatmap shows the cosine similarity between the Bregman-PC and BP parameter gradients after inference, at the last training step, averaged over three random seeds.

\section{Compute resources} 
\label{compute-res}
The experiments associated with Table 2 and Figure~\ref{fig:pc-regimes} were run on TPUs supported by Google Cloud Credits awarded through the Google TPU Research Cloud Programme (2026). All other simulations were run on a laptop CPU and took no more than a few minutes.

\section{Supplementary figures} 
\label{supp-figures}
\begin{figure}[H]
    \begin{center}
        \centerline{\includegraphics[width=0.8\textwidth]{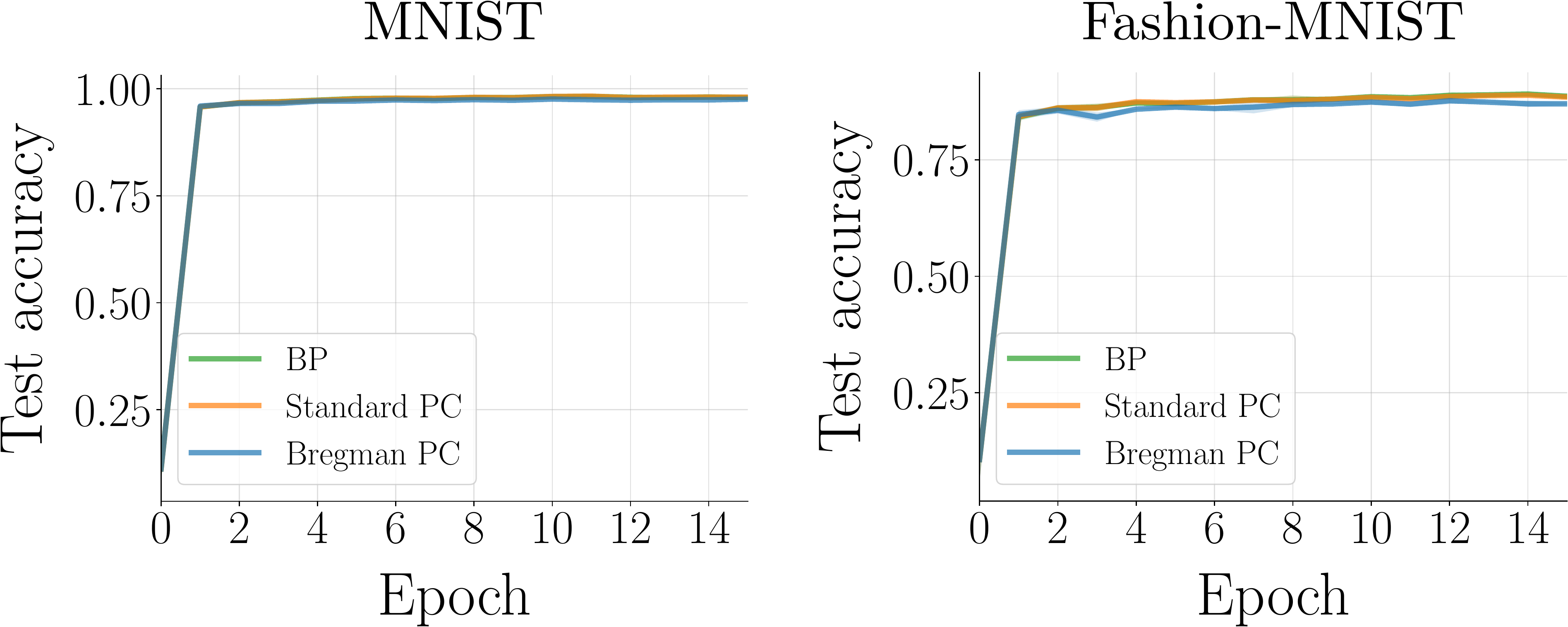}}
        \caption{\textbf{Classification learning curves on MNIST and Fashion-MNIST.} Validation classification accuracy during training for BP, standard PC, and Bregman PC on MNIST and Fashion-MNIST. Final validation accuracies are reported in Table~\ref{tab:benchmark-performance}.}
        \label{fig:classification-metrics}
    \end{center}
\end{figure}
\begin{figure}[H]
    \begin{center}
        \centerline{\includegraphics[width=\textwidth]{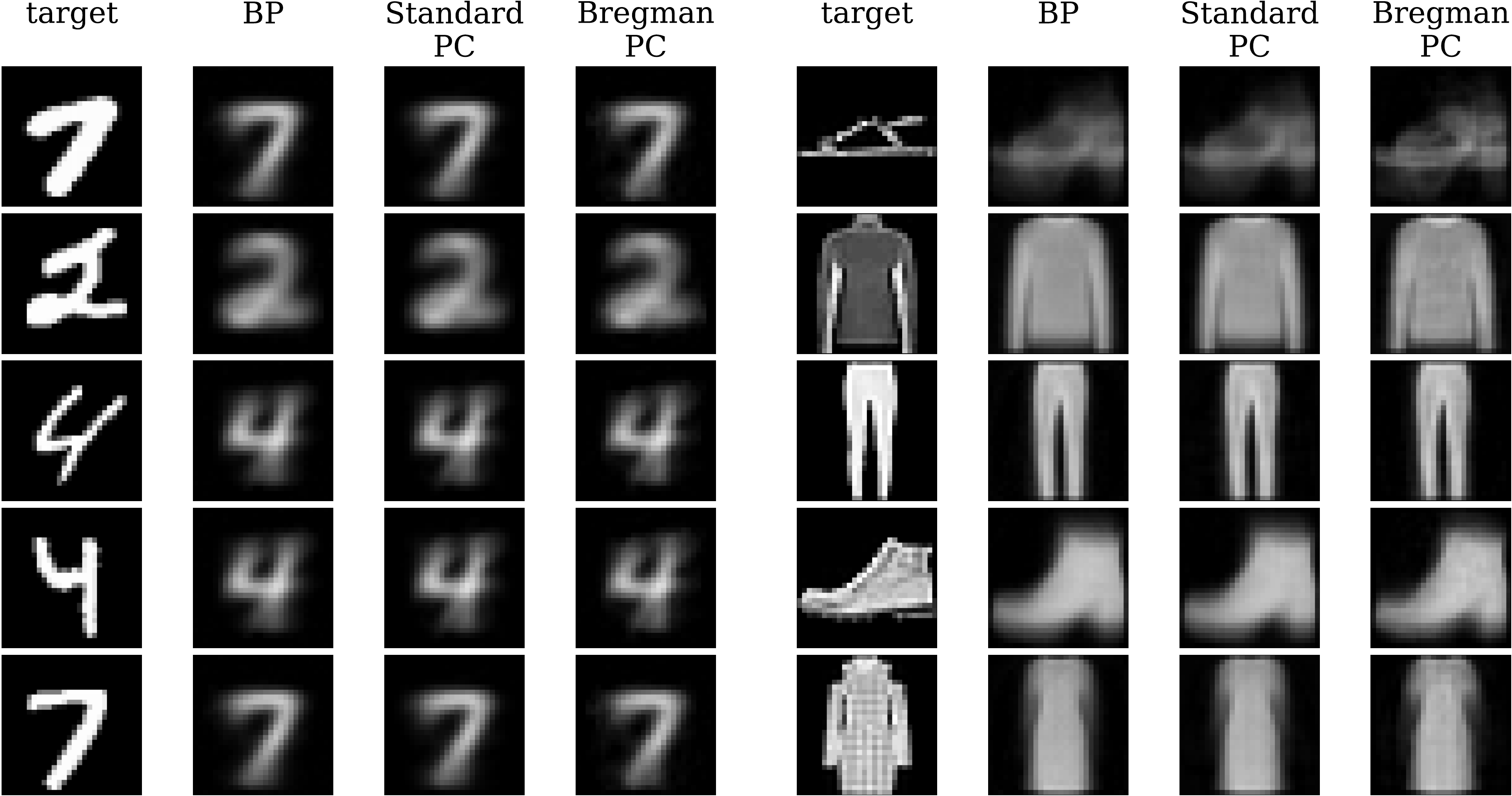}}
        \caption{\textbf{Generated samples on MNIST and Fashion-MNIST.} Representative images generated by models trained with BP, standard PC, and Bregman PC on MNIST and Fashion-MNIST. Corresponding reconstruction losses are reported in Table~\ref{tab:benchmark-performance}.}
        \label{fig:generation-examples}
    \end{center}
\end{figure}

\end{document}